\documentclass[letterpaper]{article}

\usepackage{aaai2027}

\usepackage[hyphens]{url}
\usepackage{graphicx}
\usepackage{natbib}
\usepackage{caption}
\usepackage{amsmath}
\usepackage{amssymb}
\usepackage{amsthm}

\usepackage{booktabs}
\usepackage{multirow}
\usepackage{tabularx}   
\usepackage{array}      

\usepackage{algorithm}
\usepackage{algorithmic}

\usepackage{subcaption}

\usepackage{xcolor}
\definecolor{mdeepblue}{RGB}{45,105,190}
\definecolor{mblue}{RGB}{65,180,210}
\definecolor{myellow}{RGB}{245,165,35}
\definecolor{teal}{RGB}{20,145,130}

\nocopyright

\title{
On the Reliability of Majority-Vote Pseudo-Labels
in Test-Time Reinforcement Learning
}

\author{
    Haoyu Wang\textsuperscript{\rm 1},
    Ru Wang\textsuperscript{\rm 3},
    Yuanyuan Wang\textsuperscript{\rm 1},
    Jiaxian Guo\textsuperscript{\rm 3},
    Mingming Gong\textsuperscript{\rm 1,2}
}
\affiliations{
    \textsuperscript{\rm 1}Mohamed bin Zayed University of Artificial Intelligence\\
    \textsuperscript{\rm 2}The University of Melbourne\\
    \textsuperscript{\rm 3}The University of Tokyo\\
}

\newtheorem{theorem}{Theorem}
\newtheorem{lemma}{Lemma}
\newtheorem{corollary}{Corollary}
\newtheorem{proposition}{Proposition}
\newtheorem*{restatedlemma}{Lemma}
\newtheorem*{appendixlemma}{Lemma}

\begin{document}

\maketitle

\begin{abstract}
Test-time reinforcement learning (TTRL) adapts pretrained language models to
unlabeled target-domain data using self-generated pseudo-labels, whose
reliability remains poorly understood. We analyze majority-vote supervision
for TTRL with online Direct Preference Optimization (DPO) and decompose its
error into population-majority bias and prompt-dependent finite-rollout
variation. Because population-majority bias is generally unidentifiable
without external supervision, we focus on mitigating the downstream effects
of finite-rollout variation. We propose \emph{Prompt-Conditional
Distributionally Robust DPO (PC-Dr.DPO)}, which conditions within-prompt
robust reweighting on observed vote reliability to make updates from
sampling-sensitive prompts more conservative. We further introduce a
cache-based mechanism to reduce self-supervised forgetting during online
adaptation. Across six mathematical and knowledge-intensive reasoning benchmarks and five model backbones, PC-Dr.DPO matches or outperforms the strongest TTRL baseline in 24 of 30 settings.
\end{abstract}

\section{Introduction}

Large language models (LLMs) have achieved strong performance across
reasoning, planning, and knowledge-intensive tasks through large-scale
pretraining and post-training alignment
\cite{openai2026gpt56,google2026gemini35,deepseek2026v4,
moonshot2026kimik3,qwen2026qwen37}.
However, conventional post-training is predominantly \emph{offline}: model
parameters are fixed before deployment and cannot adapt to shifts in the
test-time input distribution. This limitation is especially important in new
or specialized domains, where labeled adaptation data and reliable external
verifiers may be unavailable. It motivates test-time learning methods that
adapt LLMs using unlabeled target-domain inputs and signals generated by
the model itself.

\noindent\textbf{Test-time reinforcement learning.}
Test-time reinforcement learning (TTRL) adapts a model on unlabeled
target-domain prompts using self-generated rewards
\cite{zuo2025ttrltesttimereinforcementlearning}.
Existing approaches construct such rewards from either
\emph{certainty-based} signals, including prediction entropy and
self-certainty
\cite{prabhudesai2025rent,zhao2025intuitor},
or \emph{ensemble-based} signals, including majority-vote consistency and
agreement across alternative views of an input
\cite{zuo2025ttrltesttimereinforcementlearning,
zhang2025co,wang2025self}.
Although these methods can improve reasoning without ground-truth labels,
their success depends critically on the reliability of the resulting
self-supervision.

Despite encouraging empirical results, this reliability remains incompletely
understood. Prior work reports improvements even under imperfect or spurious
rewards
\cite{shao2025spuriousrewardsrethinkingtraining},
whereas other studies highlight risks such as memorization and contamination
\cite{wu2025reasoningmemorizationunreliableresults}.
Recent work further documents failure and collapse patterns in TTRL
\cite{he2026far}.
Nevertheless, the lack of a formal framework characterizing when
self-generated pseudo-labels are reliable limits the principled design of
test-time adaptation algorithms.

In this work, we study majority-vote pseudo-labels for TTRL with online
Direct Preference Optimization (DPO) \cite{rafailov2024directpreferenceoptimizationlanguage, pang2024iterativereasoningpreferenceoptimization, ko2024seraselfreviewingalignmentlarge}. We identify two distinct sources of
majority-vote error: (i) \emph{population-majority bias}, which occurs when
the behavior policy's population-modal answer is incorrect, and
(ii) \emph{finite-rollout variation}, which occurs when the sampled majority
differs from the population mode. The former cannot generally be identified
or corrected without external supervision. The latter instead reflects
sampling sensitivity and varies across prompts according to how decisively
the behavior policy favors its modal answer over competing answers.

\begin{figure}[t]
    \centering
    \includegraphics[width=\columnwidth]
    {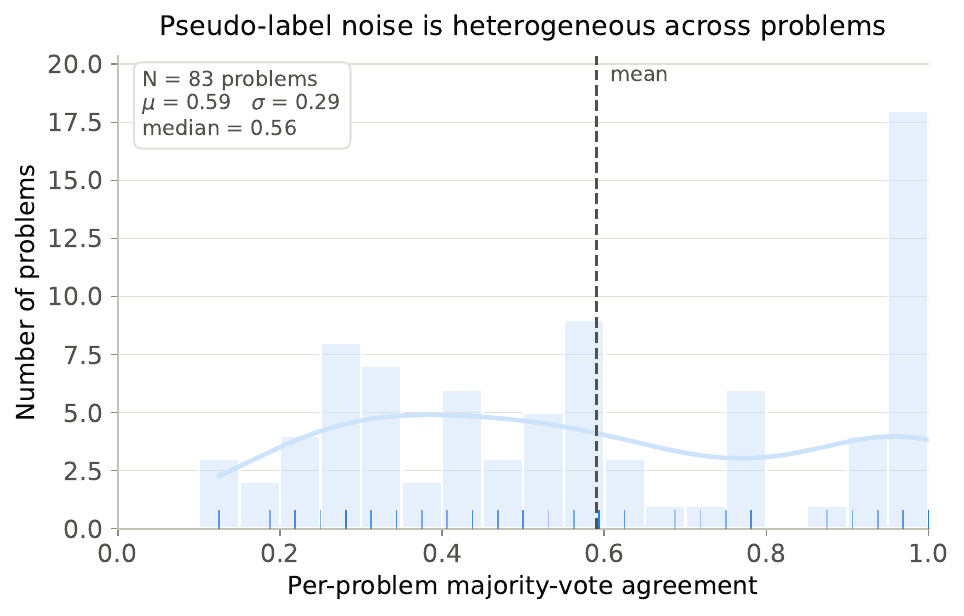}
    \caption{
    Distribution of prompt-level modal-answer agreement for
    Qwen2.5-Math-1.5B-Instruct on the 83-problem AMC benchmark.
    For each prompt $x$, we sample $m=32$ rollouts and compute the fraction producing the most
    frequent final answer. The broad distribution indicates substantial
    prompt-level heterogeneity: values near one correspond to nearly
    unanimous rollout groups, whereas smaller values indicate fragmented
    answer distributions that are more sensitive to resampling.
    }
    \label{fig:majority-fraction-dist}
\end{figure}

This prompt-level heterogeneity is important for preference construction.
Figure~\ref{fig:majority-fraction-dist} shows the distribution of the
observed modal-answer fraction for
Qwen2.5-Math-1.5B-Instruct on AMC. Although some prompts produce nearly
unanimous rollout groups, others exhibit highly fragmented answer
distributions. For the latter prompts, the selected pseudo-label is more
sensitive to the particular rollout sample and may change under resampling.

Existing noise-robust preference optimization methods typically employ either
a shared dataset-level noise model or a globally normalized robust objective
over preference pairs
\cite{mitchell2023dponoise,wu2025robustalignmentlanguagemodels}.
Consequently, they do not explicitly adapt the strength of robustness to the
prompt-dependent reliability induced by finite-rollout sampling during
test-time adaptation.

\noindent\textbf{Our approach: PC-Dr.DPO.}
To address this mismatch, we propose
\emph{Prompt-Conditional Distributionally Robust DPO (PC-Dr.DPO)}.
PC-Dr.DPO performs robust reweighting separately within each prompt's rollout
group and adapts a prompt-specific robustness temperature using observed
majority-vote reliability as a proxy for sampling sensitivity. For
low-reliability prompts, the resulting reweighting emphasizes preference
pairs already better satisfied by the current policy, yielding more
conservative updates that mitigate the downstream optimization effects of
finite-rollout errors. We further introduce a cache-based stabilization
mechanism to reduce self-supervised forgetting during long-horizon online
adaptation.

\noindent\textbf{Our contributions are threefold:}
\begin{itemize}
    \item We develop a theoretical framework that separates population-majority bias from finite-rollout variation and characterizes
    how majority-vote errors propagate into the generalization risk of models
    trained on pseudo-preferences with a DPO-style objective.

    \item We propose \emph{PC-Dr.DPO}, which uses prompt-level vote
    reliability to adapt within-prompt robust reweighting and reduce the
    influence of sampling-sensitive pseudo-labels. We further introduce a
    cache-based mechanism to mitigate self-supervised forgetting.

    \item We evaluate PC-Dr.DPO across six mathematical and
    knowledge-intensive reasoning benchmarks and five model backbones,
    matching or outperforming the strongest TTRL baseline in 24 of 30
    settings.
\end{itemize}
\section{Preliminaries}
\label{sec:prelim}

We briefly review the learning setting and the preference-optimization objectives that
our analysis and method build upon. We keep the discussion minimal and focus on the
notation used throughout the paper.

\subsection{Problem Setup and RLVR Objective}
\label{sec:setup}

Let $\pi_\theta$ denote a large language model parameterized by $\theta$, and let $\mathcal{D}$ be a dataset of question–answer pairs $(x, a)$. Given a prompt
$x$, the model generates a response trajectory
$y=(y_{t1},\ldots,y_{tn})\sim\pi_\theta(\cdot\mid x)$ autoregressively.
We consider reasoning-style responses that contain intermediate steps (e.g., CoT \cite{wang2023selfconsistencyimproveschainthought}) and a
final answer extracted by a deterministic function $\mathrm{ans}(\cdot)$.

In the standard reinforcement learning with verifiable rewards (RLVR)
setting \cite{deepseekai2025deepseekr1incentivizingreasoningcapability}, we assume access to a verifier that compares the extracted answer
with the ground-truth answer $a$, yielding the binary reward
$r(x,y;a)=\mathbf{1}\{\mathrm{ans}(y)=a\}$.
Training with verifiable rewards is commonly formulated as KL-regularized policy
optimization \cite{ouyang2022traininglanguagemodelsfollow}:

\begin{equation}
\label{eq:rlvr_obj}
\max_{\pi_\theta}\!
\mathbb{E}_{x \sim \mathcal{D}}\!
\Big[
\mathbb{E}_{y \sim \pi_\theta(\cdot \mid x)}\big[r(x, y)\big]
- \tau\, D_{\mathrm{KL}}\!\big(\pi_\theta(\cdot \! \mid x)\,\|\,\pi_{\mathrm{ref}}(\cdot \!\mid x)\big)
\Big].
\end{equation}

where $\pi_{\mathrm{ref}}$ is a reference policy (typically an SFT model) and $\tau>0$
controls the strength of the KL regularization. 
In practice, several methods have been proposed to optimize~\eqref{eq:rlvr_obj}, including PPO~\cite{schulman2017proximalpolicyoptimizationalgorithms},
RLOO~\cite{ahmadian2024basicsrevisitingreinforcestyle}, REMAX~\cite{li2024remaxsimpleeffectiveefficient}, REINFORCE++~\cite{hu2025reinforceefficientrlhfalgorithm}, and, most prominently, the GRPO family algorithms~\cite{shao2024deepseekmathpushinglimitsmathematical,yu2025dapoopensourcellmreinforcement,zheng2025groupsequencepolicyoptimization}.

\paragraph{Test-time self-reward via majority voting.}
Unlike standard RLVR, the TTRL setting considered here assumes neither
ground-truth answers nor an external verifier at adaptation time. Instead,
the model constructs self-supervision from its own rollouts. For each prompt
$x$, the behavior policy samples $m$ independent responses
$\{y_i\}_{i=1}^{m}$ with $y_i\sim\pi_b(\cdot\mid x)$ and forms the
pseudo-label
$\widehat a_{\mathrm{mv}}(x)
=
\operatorname*{arg\,max}_{a'}
\sum_{i=1}^{m}
\mathbf{1}\{\operatorname{ans}(y_i)=a'\}$
by majority voting over their extracted answers. This pseudo-label is used
either to assign pseudo-rewards or to construct pseudo-preference pairs,
treating matching rollouts as correct under the pseudo-label and the
remaining rollouts as incorrect. Section~\ref{sec:analysis} analyzes the
reliability of this supervision under a DPO-style objective.

\subsection{Direct Preference Optimization}
\label{sec:dpo}

Rather than performing explicit policy-gradient optimization, Direct
Preference Optimization (DPO)
\cite{rafailov2024directpreferenceoptimizationlanguage}
provides a supervised objective for optimizing the KL-regularized reward
maximization problem using winner--loser preference pairs.

Let $\mathcal D_{\mathrm{pref}}$ denote a dataset of preference tuples
$(x,y_w,y_l)$, where $y_w\succ y_l$ indicates that $y_w$ is preferred to
$y_l$ under prompt $x$. DPO minimizes
\begin{equation}
\label{eq:dpo_loss}
\mathcal L_{\mathrm{DPO}}(\theta)
=
-\mathbb E_{(x,y_w,y_l)\sim\mathcal D_{\mathrm{pref}}}
\left[
h_\theta(x,y_w,y_l)
\right],
\end{equation}
where the per-pair log-preference score is
\begin{equation}
\label{eq:dpo_score}
h_\theta(x,y_w,y_l)
=
\log\sigma\!\left(
\tau g_\theta(x,y_w,y_l)
\right),
\end{equation}
and
\begin{equation}
\label{eq:dpo_margin}
g_\theta(x,y_w,y_l)
=
\log
\frac{\pi_\theta(y_w\mid x)}
     {\pi_{\mathrm{ref}}(y_w\mid x)}
-
\log
\frac{\pi_\theta(y_l\mid x)}
     {\pi_{\mathrm{ref}}(y_l\mid x)}.
\end{equation}
Here, $\sigma(\cdot)$ denotes the sigmoid function. We refer to
$g_\theta$ as the DPO margin and to $h_\theta$ as the corresponding
log-preference score.

Standard DPO operates on a fixed offline preference dataset. Online variants
instead generate preference data iteratively from a behavior policy
\cite{pang2024iterativereasoningpreferenceoptimization}.

\subsection{Distributionally Robust DPO}
\label{sec:drdpo}

Dr.DPO~\cite{wu2025robustalignmentlanguagemodels} applies KL-based
distributionally robust reweighting to reduce the influence of pairs that are
inconsistent with the current policy-relative preference ordering. Let
$s=(x,y_w,y_l)$ denote a preference sample and let $p$ be the empirical
distribution over the preference dataset $\mathcal D_{\mathrm{pref}}$.
Motivated by the dual form of a KL-constrained distributionally robust
objective, Dr.DPO uses the fixed-temperature loss
\begin{equation}
\label{eq:drdpo_loss}
\mathcal L_{\mathrm{DrDPO}}(\theta)
=
-\beta
\log
\mathbb E_{s\sim p}
\left[
\exp\!\left(
\frac{h_\theta(s)}{\beta}
\right)
\right],
\qquad
\beta>0,
\end{equation}
where $h_\theta$ is the DPO log-preference score defined in
Equation~\eqref{eq:dpo_score}.

Define the current pairwise preference probability as
\[
\nu_\theta(s)
:=
\sigma\!\left(\tau g_\theta(s)\right),
\]
so that $h_\theta(s)=\log\nu_\theta(s)$. The objective in
Equation~\eqref{eq:drdpo_loss} induces the normalized exponential density
ratio
\begin{equation}
\label{eq:drdpo_weight}
w_\beta(s)
=
\frac{
\exp\!\left(h_\theta(s)/\beta\right)
}{
\mathbb E_{s'\sim p}
\left[
\exp\!\left(h_\theta(s')/\beta\right)
\right]
},
\end{equation}
and hence
\[
w_\beta(s)
\propto
\nu_\theta(s)^{1/\beta}.
\]
The corresponding gradient is
\begin{equation}
\label{eq:drdpo_gradient}
\nabla_\theta
\mathcal L_{\mathrm{DrDPO}}(\theta)
=
-\tau\,
\mathbb E_{s\sim p}
\left[
w_\beta(s)
\bigl(1-\nu_\theta(s)\bigr)
\nabla_\theta g_\theta(s)
\right].
\end{equation}
Smaller values of $\beta$ produce a sharper exponential tilt toward pairs
with larger current preference probabilities. The KL-DRO derivation and a
detailed analysis of the effect of $\beta$ on the resulting gradient are
provided in Appendix~A.1.
\section{Analysis of Majority-Vote Pseudo-Labels}
\label{sec:analysis}

We first analyze the reliability of majority-vote pseudo-labels in test-time
reinforcement learning by separating two sources of error:
\emph{population-majority bias}, where the modal answer under the behavior
policy is incorrect, and \emph{finite-rollout variation}, where a sampled
majority differs from that population mode.

Throughout this section, we condition on a fixed behavior policy $\pi_b$ and
adopt an i.i.d.\ preference-pair abstraction. Each sample
$s=(x,y_w,y_l)$ is independently constructed from $m$ rollouts for a sampled
prompt using majority-vote preference labeling. The analysis applies
separately to each online update, with $\pi_b$ fixed to the rollout-generating
policy at that update. Let $\mathcal P$ denote the resulting distribution over
preference samples.

\subsection{Generalization under Approximate Preference Labels}
\label{sec:risk-decomp}

Let $\mathcal S$ be the preference-sample space and
$\mathcal F\subseteq\{f:\mathcal S\to\mathbb R\}$ a class of pairwise
predictors. For $s=(x,y_w,y_l)$, let $z\in\{0,1\}$ denote the latent true
preference label, with $z=1$ corresponding to $y_w\succ y_l$. Since the pair
is ordered by the pseudo-labeling procedure, $z$ indicates whether the
constructed ordering agrees with the true preference. We write
$p^*(s):=(p^*(z=1\mid s),p^*(z=0\mid s))^\top\in\Delta_2$.

For DPO, the predictor is instantiated by the policy-relative margin
\begin{equation}
\label{eq:dpo-margin-analysis}
f_\theta(s)
:=
\tau\!\left[
\log\frac{\pi_\theta(y_w\mid x)}
          {\pi_{\mathrm{ref}}(y_w\mid x)}
-
\log\frac{\pi_\theta(y_l\mid x)}
          {\pi_{\mathrm{ref}}(y_l\mid x)}
\right],
\end{equation}
where $\tau>0$ is the inverse-temperature parameter. The corresponding
binary logistic loss is
$\ell_{\mathrm{DPO}}(z,f)
=-z\log\sigma(f)-(1-z)\log\sigma(-f)$.
Because the generalization result below requires a loss bounded in $[0,1]$,
we assume $|f(s)|\le M$ for every $f\in\mathcal F$ and $s\in\mathcal S$,
and define
$\ell(z,f):=\ell_{\mathrm{DPO}}(z,f)/\log(1+e^M)\in[0,1]$.
This positive rescaling preserves empirical minimizers.

Let
$\ell(f(s)):=(\ell(1,f(s)),\ell(0,f(s)))^\top$. The true population risk is
defined using $p^*(s)$. Since $p^*(s)$ is unavailable at test time, empirical
optimization instead uses the pseudo-preference distribution
$\widehat p(s)\in\Delta_2$:
\begin{equation}
\label{eq:preference-risks}
\begin{aligned}
R(f)
&:=
\mathbb E_{s\sim\mathcal P}
\!\left[p^*(s)^\top\ell(f(s))\right],\\
\widetilde R(f;S)
&:=
\frac1N\sum_{n=1}^N
\widehat p(s_n)^\top\ell(f(s_n)),
\end{aligned}
\end{equation}
where $S=\{s_n\}_{n=1}^N\sim\mathcal P^N$.

Define the induced loss class and its covering complexity by
\begin{equation}
\label{eq:approx-induced-class}
\begin{aligned}
\widetilde{\mathcal H}
&:=
\left\{
s\mapsto\widehat p(s)^\top\ell(f(s))
:\ f\in\mathcal F
\right\},\\
\widetilde{\mathcal M}_N
&:=
\mathcal N_\infty
\left(
\tfrac1N,\widetilde{\mathcal H},2N
\right).
\end{aligned}
\end{equation}
Following \citet{menon2021statistical},
$\mathcal N_\infty(\epsilon,\mathcal H,n)$ denotes the uniform
$\ell_\infty$ covering number of $\mathcal H$ over samples of size $n$.
For brevity, define the sample-dependent generalization term
\begin{equation}
\label{eq:generalization-term}
\mathcal G_N(f,\delta;S)
:=
O\!\left(
\sqrt{
\frac{
\widetilde{\mathbb V}_N(f;S)
\log(\widetilde{\mathcal M}_N/\delta)}
{N}}
+
\frac{\log(\widetilde{\mathcal M}_N/\delta)}{N}
\right),
\end{equation}
where $\widetilde{\mathbb V}_N(f;S)$ is the empirical variance of
$\{\widehat p(s_n)^\top\ell(f(s_n))\}_{n=1}^N$.

\begin{theorem}[Approximate preference distillation]
\label{thm:approx-distillation}
For any $\delta\in(0,1)$, with probability at least $1-\delta$ over
$S\sim\mathcal P^N$, every $f\in\mathcal F$ satisfies
\begin{equation}
\label{eq:approx-bound}
R(f)
\le
\widetilde R(f;S)
+
\frac12\,
\mathbb E_{s\sim\mathcal P}
\!\left[
\left\|
\widehat p(s)-p^*(s)
\right\|_1
\right]
+
\mathcal G_N(f,\delta;S).
\end{equation}
\end{theorem}

The second term in Equation~\eqref{eq:approx-bound} is the expected total-variation distance between the pseudo-preference and true conditional preference distributions. Unlike $\mathcal G_N(f,\delta;S)$, this mismatch need not vanish as $N$ increases and therefore contributes persistently to the bound when pseudo-label error is systematic. The complete proof is given in Appendix~A.2.

\subsection{Majority-Vote Error Decomposition}
\label{sec:mv-decomposition}

To expose the central mechanism, we consider a binary answer space and assume
deterministic ground-truth preferences. For any pair with distinct final
answers, the true preference distribution $p^*(s)$ assigns probability one
to the response whose extracted answer equals the ground-truth answer
$a^\star(x)$. Define the per-rollout correctness probability and the
population-majority answer under the behavior policy $\pi_b$ by
\begin{equation}
\label{eq:rho-pop-majority}
\begin{aligned}
\rho_b(x)
&:=
\Pr_{\pi_b}\!\left(
\operatorname{ans}(y)=a^\star(x)\mid x
\right),\\
a_b^\infty(x)
&:=
\arg\max_a
\Pr_{\pi_b}\!\left(
\operatorname{ans}(y)=a\mid x
\right).
\end{aligned}
\end{equation}
In the binary setting, $a_b^\infty(x)=a^\star(x)$ when
$\rho_b(x)>1/2$, whereas the population majority is incorrect when
$\rho_b(x)<1/2$. We assume that the rollout count $m$ is odd, so the sampled
majority is unique, and adopt a fixed tie-breaking rule for the population
boundary case $\rho_b(x)=1/2$.

Given rollouts
$y_1,\ldots,y_m\overset{\mathrm{i.i.d.}}{\sim}
\pi_b(\cdot\mid x)$, define the sampled-majority answer by
\begin{equation}
\label{eq:sample-majority}
\widehat a_{m,b}(x)
:=
\arg\max_a
\sum_{i=1}^m
\mathbf 1\!\left\{
\operatorname{ans}(y_i)=a
\right\}.
\end{equation}
A response agreeing with $\widehat a_{m,b}(x)$ is treated as the winner,
whereas a response containing the alternative answer is treated as the
loser. Thus, the pseudo-preference distribution assigns probability one to
the constructed ordering, so $\widehat p(s)=(1,0)^\top$.

Define the conditional majority-vote error probability by
\begin{equation}
\label{eq:majority-error}
U_{m,b}(x)
:=
\Pr\!\left(
\widehat a_{m,b}(x)\neq a^\star(x)
\mid x
\right),
\end{equation}
Under the binary deterministic-preference abstraction, the expected
total-variation term in Theorem~\ref{thm:approx-distillation} equals
$\mathbb E_x[U_{m,b}(x)]$, as shown in Appendix~A.3. Consequently, with the
same probability $1-\delta$, every $f\in\mathcal F$ satisfies
\begin{equation}
\label{eq:mv-generalization}
R(f)
\le
\widetilde R(f;S)
+
\mathbb E_x\!\left[
U_{m,b}(x)
\right]
+
\mathcal G_N(f,\delta;S),
\end{equation}
where the expectation is taken with respect to the prompt marginal induced
by $\mathcal P$.

\subsection{Population Bias and Finite-Rollout Variation}
\label{sec:majority-concentration}

We now decompose the majority-vote error into population-majority
bias and finite-rollout variation:
\begin{equation}
\label{eq:bias-variation-defs}
\begin{aligned}
B_b(x)
&:=
\mathbf 1
\left\{
a_b^\infty(x)\neq a^\star(x)
\right\},\\
V_{m,b}(x)
&:=
\Pr\!\left(
\widehat a_{m,b}(x)\neq a_b^\infty(x)
\mid x
\right).
\end{aligned}
\end{equation}

\begin{proposition}[Exact majority-error decomposition]
\label{prop:majority-decomposition}
Under the binary-answer abstraction, for every prompt $x$,
\begin{align}
U_{m,b}(x)
&=
\bigl(1-B_b(x)\bigr)V_{m,b}(x)
+
B_b(x)\bigl(1-V_{m,b}(x)\bigr)
\nonumber\\
&=
B_b(x)
+
\bigl(1-2B_b(x)\bigr)V_{m,b}(x).
\label{eq:compact-decomposition}
\end{align}
\end{proposition}

\begin{lemma}[Majority-vote concentration]
\label{lem:majority-concentration}
For every prompt $x$ with $\rho_b(x)\neq 1/2$,
\begin{equation}
\label{eq:pointwise-concentration}
V_{m,b}(x)
\le
\exp\!\left(
-2m\left|\rho_b(x)-\tfrac12\right|^2
\right).
\end{equation}
\end{lemma}

Combining Proposition~\ref{prop:majority-decomposition} and
Lemma~\ref{lem:majority-concentration}, for any $\gamma>0$,
\begin{equation}
\label{eq:majority-gap-consequences}
\begin{aligned}
\rho_b(x)\ge\tfrac12+\gamma
&\Longrightarrow
U_{m,b}(x)\le e^{-2m\gamma^2},\\
\rho_b(x)\le\tfrac12-\gamma
&\Longrightarrow
U_{m,b}(x)\ge1-e^{-2m\gamma^2}.
\end{aligned}
\end{equation}

\begin{corollary}[Generalization under majority-vote supervision]
\label{cor:bias-variation-separation}
Suppose that
$\left|\rho_b(x)-1/2\right|\ge\gamma>0$ almost surely, and let
\[
\alpha_b
:=
\Pr_x\!\left(
\rho_b(x)<\tfrac12
\right)
=
\mathbb E_x[B_b(x)].
\]
Then
\begin{equation}
\label{eq:error-to-bias}
\left|
\mathbb E_x[U_{m,b}(x)]-\alpha_b
\right|
\le
e^{-2m\gamma^2}.
\end{equation}
Consequently, with probability at least $1-\delta$ over
$S\sim\mathcal P^N$, every $f\in\mathcal F$ satisfies
\begin{equation}
\label{eq:final-generalization-bound}
R(f)
\le
\widetilde R(f;S)
+
\alpha_b
+
e^{-2m\gamma^2}
+
\mathcal G_N(f,\delta;S).
\end{equation}
\end{corollary}

Proofs are provided in Appendix~A.4--A.5.

\paragraph{Interpretation.}
For a fixed behavior policy, $\alpha_b$ captures population-majority bias
and does not decrease with the rollout count $m$. By contrast, finite-rollout variation
decays exponentially in $m$. Increasing $m$ can therefore stabilize the
sampled majority but cannot correct an incorrect population majority.
Appendix~C.1 empirically illustrates this distinction using
Qwen2.5-Math-1.5B-Instruct on AMC with $m\in\{16,32,64\}$.

\section{Our Method: Prompt-Conditional Dr.DPO}
\label{sec:method}

The preceding analysis separates majority-vote error into
population-majority bias and finite-rollout variation under a simplified
binary-answer abstraction. Without external supervision, the former is
generally unidentifiable: an incorrect but decisive population majority may
still appear highly reliable. We therefore focus on mitigating the downstream
optimization effects of finite-rollout variation. Our method uses observed
vote concentration as a proxy for sampling sensitivity and adjusts the
strength of robust reweighting separately for each prompt.

\subsection{Prompt-Conditional Robust Objective}
\label{sec:pcdro}

Let $p$ denote the empirical distribution over preference samples
$s=(x,y_w,y_l)$. Standard Dr.DPO applies a globally normalized exponential
tilt with a shared temperature across all preference pairs. To account for
prompt-dependent reliability, we instead consider the conditional
uncertainty set
\[
\mathcal U
=
\left\{
q:
q(x)=p(x),\
\mathrm{KL}\!\left(
q(\cdot\mid x)\,\|\,p(\cdot\mid x)
\right)
\le \eta_x,\ \forall x
\right\}.
\]
This set keeps the prompt marginal fixed while permitting prompt-specific
reweighting of the conditional pair distribution. Given the DPO score
$h_\theta$ defined in Equation~\eqref{eq:dpo_score}, the corresponding
robust score is
\begin{equation}
\label{eq:pcdro-primal}
\sup_{q\in\mathcal U}
\mathbb E_{s\sim q}
\left[
h_\theta(s)
\right].
\end{equation}

Because $q(x)=p(x)$, the inner optimization decomposes across prompts:
\begin{equation}
\label{eq:pcdro-decomposition}
\mathbb E_{x\sim p(x)}
\left[
\sup_{\substack{q_x:\,
\mathrm{KL}(q_x\|p(\cdot\mid x))\le\eta_x}}
\mathbb E_{s\sim q_x}
\left[
h_\theta(s)
\right]
\right].
\end{equation}
Applying KL-ball duality to each conditional problem gives
\begin{equation}
\label{eq:pcdro-dual}
\mathbb E_{x\sim p(x)}
\left[
\inf_{\beta_x>0}
\left\{
\beta_x\eta_x
+
\beta_x
\log
\mathbb E_{s\sim p(\cdot\mid x)}
\left[
\exp\!\left(
\frac{h_\theta(s)}{\beta_x}
\right)
\right]
\right\}
\right].
\end{equation}

Let $\beta_x=\beta_x^\star(\eta_x)$ denote the prompt-specific dual
temperature associated with the prescribed radius $\eta_x$. Following the
fixed-temperature formulation of Dr.DPO, we hold $\beta_x$ fixed during the
policy update and omit the additive term $\beta_x\eta_x$, which is
independent of $\theta$. This yields the prompt-conditional score
\begin{equation}
\label{eq:pcdro}
\mathbb E_{x\sim p(x)}
\left[
\beta_x
\log
\mathbb E_{s\sim p(\cdot\mid x)}
\left[
\exp\!\left(
\frac{h_\theta(s)}{\beta_x}
\right)
\right]
\right].
\end{equation}

The temperature $\beta_x$ controls the sharpness of the conditional
exponential tilt. Smaller values concentrate weight on pairs with larger
current DPO scores, whereas larger values produce weights closer to uniform
within the prompt group.

\paragraph{Finite-sample objective.}
At adaptation step $t$, rollouts are sampled from
$\pi_{t-1}$, which also serves as the reference policy:
$\pi_{\mathrm{ref}}=\pi_{t-1}$. For each prompt $x$, let
\[
P_t(x)=\{s_{xi}\}_{i=1}^{N_x},
\qquad
s_{xi}=(x,y_{w,i},y_{l,i}),
\]
denote the resulting preference pairs. The prompt-conditional objective is
\begin{equation}
\label{eq:pcdrdpo-loss}
\mathcal L_{\mathrm{PC\text{-}DrDPO}}(\theta;x)
=
-\beta_x
\log
\left[
\frac{1}{N_x}
\sum_{i=1}^{N_x}
\exp\!\left(
\frac{h_\theta(s_{xi})}{\beta_x}
\right)
\right].
\end{equation}
Unlike standard Dr.DPO, which normalizes globally over preference pairs,
Equation~\eqref{eq:pcdrdpo-loss} normalizes within each prompt group. Let
\[
\nu_{xi}
:=
\sigma\!\left(\tau g_\theta(s_{xi})\right),
\qquad
h_\theta(s_{xi})=\log\nu_{xi}.
\]
The induced weights and gradient are
\begin{align}
\overline w_{xi}(\beta_x)
&=
\frac{\nu_{xi}^{1/\beta_x}}
{\sum_{k=1}^{N_x}\nu_{xk}^{1/\beta_x}},
\label{eq:pcdrdpo-weights}\\
\nabla_\theta
\mathcal L_{\mathrm{PC\text{-}DrDPO}}(\theta;x)
&=
-\tau
\sum_{i=1}^{N_x}
\overline w_{xi}(\beta_x)
(1-\nu_{xi})
\nabla_\theta g_\theta(s_{xi}).
\label{eq:pcdrdpo-gradient}
\end{align}
As $\beta_x\to0^+$, the objective approaches
$-\max_i h_\theta(s_{xi})$ and the weights concentrate on the
highest-scoring pairs; as $\beta_x\to\infty$, the weights become uniform
over $P_t(x)$.

\subsection{Reliability-Conditioned Temperature}
\label{sec:prompt-temperature}

For each prompt $x$, let
$\{y_i\}_{i=1}^{m}\overset{\mathrm{i.i.d.}}{\sim}
\pi_{t-1}(\cdot\mid x)$, and let $\widehat a_{m,t-1}(x)$ denote the
sampled modal-answer. We define
\begin{equation}
\label{eq:majority-reliability}
\kappa(x)
:=
\frac{1}{m}
\sum_{i=1}^{m}
\mathbf 1
\left\{
\operatorname{ans}(y_i)
=
\widehat a_{m,t-1}(x)
\right\}.
\end{equation}
The modal-answer fraction $\kappa(x)$ does not determine whether the resulting
pseudo-label is correct. Instead, it serves as an observable proxy for
finite-rollout stability.

We set
\begin{equation}
\label{eq:beta-def}
\beta_x
=
\beta_{\mathrm{base}}\kappa(x),
\qquad
\beta_{\mathrm{base}}>0.
\end{equation}
Thus, weaker observed majorities receive smaller temperatures and sharper
within-prompt reweighting, whereas stronger majorities receive smoother
weights closer to uniform averaging. The temperature is computed from the
rollout group and held fixed during policy differentiation.

\begin{proposition}[Effect of prompt-conditional temperature]
\label{prop:temperature-effect}
Fix $\theta$ and a prompt group
$P_t(x)=\{s_{xi}\}_{i=1}^{N_x}$. Let
$\nu_{xi}:=\nu_\theta(s_{xi})$ and define
\begin{equation}
\label{eq:prompt-residual-factor}
A_x(\beta)
:=
\sum_{i=1}^{N_x}
\overline w_{xi}(\beta)(1-\nu_{xi}).
\end{equation}
Then $A_x(\beta)$ is nondecreasing in $\beta$. Moreover, if
$\|\nabla_\theta g_\theta(s_{xi})\|\le G$ for all $i$, then
\begin{equation}
\label{eq:prompt-gradient-bound-method}
\left\|
\nabla_\theta
\mathcal L_{\mathrm{PC\text{-}DrDPO}}(\theta;x)
\right\|
\le
\tau G A_x(\beta_x).
\end{equation}
Consequently, with the current pairwise scores held fixed, decreasing
$\beta_x$ decreases this upper bound on the prompt-level update magnitude.
\end{proposition}

The proof is provided in Appendix~A.1. Intuitively, a smaller temperature
places greater weight on pairs with larger $\nu_{xi}$, for which the
residual factors $1-\nu_{xi}$ are smaller. This moderates the gradient-norm
upper bound for prompts whose sampled majorities appear less stable.

\subsection{Cache-Based Stabilization Against Self-Forgetting}
\label{sec:cache}

Some prompts become degenerate during online adaptation: all sampled
rollouts produce the same final answer, so no valid winner--loser pair can
be constructed and $P_t(x)=\emptyset$. We exclude such prompts from the preference update.

When a prompt first becomes degenerate, we cache its rollout completions,
denoted by $\mathcal C(x)$. Because all cached completions share the same
final answer, we denote this answer by $a_{\mathcal C}(x)$. If the prompt is
encountered again and its new sampled-majority answer disagrees with
$a_{\mathcal C}(x)$, we discard the newly generated pseudo-preference pairs
and instead apply the cache loss
\begin{equation}
\label{eq:cache-loss}
\mathcal L_{\mathrm{cache}}(\theta;x)
=
-\frac{1}{|\mathcal C(x)|}
\sum_{y\in\mathcal C(x)}
\log\pi_\theta(y\mid x).
\end{equation}
This replay-style safeguard discourages a later conflicting pseudo-label
from overwriting previously stable self-supervision.

\subsection{Overall Objective and Update Rule}
\label{sec:overall}

For a mini-batch of prompts $\mathcal B_t$, let
$\mathcal B_{\mathrm{cache}}$ denote the prompts that trigger the cache rule,
and define
\[
\mathcal B_{\mathrm{norm}}
:=
\left\{
x\in
\mathcal B_t\setminus\mathcal B_{\mathrm{cache}}
:
P_t(x)\neq\emptyset
\right\}.
\]
The complete update objective is
\begin{equation}
\label{eq:final-obj}
\begin{aligned}
\mathcal L_t(\theta)
=\;&
\frac{1}{|\mathcal B_{\mathrm{norm}}|}
\sum_{x\in\mathcal B_{\mathrm{norm}}}
\mathcal L_{\mathrm{PC\text{-}DrDPO}}(\theta;x)\\
&+
\mu \cdot
\frac{1}{|\mathcal B_{\mathrm{cache}}|}
\sum_{x\in\mathcal B_{\mathrm{cache}}}
\mathcal L_{\mathrm{cache}}(\theta;x),
\end{aligned}
\end{equation}
where $\mu\ge 0$ controls the relative weight of the cache loss.
(When $\mathcal B_{\mathrm{cache}}=\emptyset$, the second term is omitted.)

At each adaptation step, we perform one or more gradient updates on
Equation~\eqref{eq:final-obj} and set the resulting policy as $\pi_t$.
Using $\pi_{\mathrm{ref}}=\pi_{t-1}$ keeps each DPO update local to the
behavior policy that generated the corresponding rollout data.

\paragraph{Trajectory-score-based answer selection.}
Majority voting over discrete answers ignores within-answer heterogeneity.
To incorporate the model's internal preference signal, we compute an
\emph{implicit DPO score} for each rollout using successive-policy
likelihood ratios. At update step $t\ge2$, we treat $\pi_{t-2}$ as a lagged
reference and define
\begin{equation}
\label{eq:implicit_reward}
r_t(y\mid x)
:=
\log
\frac{\pi_{t-1}(y\mid x)}
     {\pi_{t-2}(y\mid x)}.
\end{equation}
We aggregate these scores by extracted final answer and select
$\widehat a^{\,\mathrm{score}}_{m,t-1}(x)
=
\operatorname*{arg\,max}_{a}
\sum_{i:\operatorname{ans}(y_i)=a}
r_t(y_i\mid x)$,
with ties resolved by a fixed rule.
Rollouts producing the selected answer
are treated as winners, whereas those producing other answers are treated
as losers when constructing $P_t(x)$. Frequency-based majority voting is
recovered by assigning every trajectory unit score. At the first adaptation
step, when $\pi_{t-2}$ is unavailable, we use frequency-based selection.
This rule affects only preference construction and leaves
Equation~\eqref{eq:final-obj} unchanged.

\begin{algorithm}
\caption{PC-Dr.DPO (Test-Time Adaptation)}
\label{alg:pcdrdpo}
\small
\begin{algorithmic}[1]
\REQUIRE Initial policy $\pi_0$; rollout size $m$; base temperature $\beta_{\mathrm{base}}$; cache weight $\mu$.
\STATE Initialize cache $\mathcal{C}\leftarrow\emptyset$.
\FOR{$t=1,2,\dots,T$}
  \STATE Sample mini-batch of prompts $\mathcal{B}\sim p(x)$.
  \FORALL{$x\in\mathcal{B}$}
    \STATE Sample rollouts $\{y_i\}_{i=1}^m \sim \pi_{t-1}(\cdot\mid x)$.
   \STATE Determine the sampled pseudo-answer $\widehat a(x)$.
    \STATE Construct pair set $P_t(x)$ from winners vs.\ losers.
    \STATE Compute majority fraction $\kappa(x)=n_{\mathrm{maj}}(x)/m$ and set $\beta_x=\beta_{\mathrm{base}}\kappa(x)$.
    \IF{$P_t(x)=\emptyset$ (degenerate)}
      \STATE Cache $\mathcal{C}(x)\leftarrow$ unique rollouts for $x$ (if not already cached).
    \ENDIF
  \ENDFOR
  \STATE Optimize $\theta_t$ by minimizing~\eqref{eq:final-obj}.
\ENDFOR
\RETURN Adapted policy $\pi_T$.
\end{algorithmic}
\end{algorithm}
\section{Experimental Setup}

\paragraph{Models.}
We evaluate PC-Dr.DPO across multiple architectures and model scales,
including Qwen2.5-1.5B/3B/7B-Instruct and
Qwen2.5-Math-1.5B-Instruct~\cite{qwen2025qwen25technicalreport}. We also
evaluate Llama-3.2-3B-Instruct \cite{meta2024llama32} to assess whether the method generalizes
across model families. We use instruction-tuned variants throughout to
avoid chat-template mismatches under distribution shift, ensuring that
performance differences primarily reflect test-time adaptation rather than
formatting artifacts.

\begin{table}[t]
\centering
\caption{Test-time adaptation accuracy (\%) under Avg@32 evaluation.
The best result among adaptation methods is \textbf{bolded}.
$\ast$ denotes training collapse.}
\label{tab:main-results}
\vspace{-1mm}

\begingroup
\fontsize{6.2pt}{7.0pt}\selectfont
\setlength{\tabcolsep}{1.5pt}
\renewcommand{\arraystretch}{0.90}
\setlength{\aboverulesep}{0.4pt}
\setlength{\belowrulesep}{0.4pt}
\setlength{\cmidrulesep}{0.3pt}

\begin{tabular*}{\columnwidth}{
@{\extracolsep{\fill}}lcccccc@{}
}
\toprule
\multirow{2}{*}{Method}
& \multicolumn{4}{c}{Math Reasoning}
& \multicolumn{2}{c}{Knowledge Reasoning} \\
\cmidrule(lr){2-5}
\cmidrule(l){6-7}
& A24 & A25 & AMC & M500 & GPQA-D & MMLU-M \\
\midrule

\multicolumn{7}{c}{\bfseries\itshape Qwen2.5-1.5B-Instruct} \\
\addlinespace[0.5pt]
Base
& 2.9 & 0.8 & 22.0 & 50.7 & 24.1 & 39.5 \\
RENT
& 0.0 & $\ast$ & 19.2 & 55.7 & $\ast$ & 5.2 \\
TTRL
& \textbf{3.3} & $\ast$ & \textbf{33.6} & 62.3
& 23.5 & 10.9 \\
Co-Reward
& \textbf{3.3} & $\ast$ & 31.4 & 59.6 & 23.3 & 10.8 \\
PC-Dr.DPO
& \textbf{3.3} & $\ast$ & 32.3 & \textbf{62.9}
& \textbf{28.8} & \textbf{45.3} \\
\midrule

\multicolumn{7}{c}{\bfseries\itshape Qwen2.5-3B-Instruct} \\
\addlinespace[0.5pt]
Base
& 6.9 & 2.4 & 34.3 & 63.5 & 29.6 & 53.6 \\
RENT
& $\ast$ & $\ast$ & $\ast$ & 56.5 & 7.8 & 46.4 \\
TTRL
& \textbf{16.7} & \textbf{3.3} & 44.8
& 70.9 & \textbf{28.8} & 46.0 \\
Co-Reward
& 13.1 & \textbf{3.3} & 41.0 & 69.6
& \textbf{28.8} & $\ast$ \\
PC-Dr.DPO
& 13.3 & \textbf{3.3} & \textbf{45.8}
& \textbf{73.3} & \textbf{28.8} & \textbf{64.7} \\
\midrule

\multicolumn{7}{c}{\bfseries\itshape Qwen2.5-7B-Instruct} \\
\addlinespace[0.5pt]
Base
& 12.2 & 6.1 & 43.1 & 74.0 & 34.3 & 72.4 \\
RENT
& 10.0 & 16.7 & 47.6 & $\ast$ & 33.5 & $\ast$ \\
TTRL
& \textbf{20.0} & \textbf{17.4} & 52.7
& 79.5 & \textbf{36.9} & 76.1 \\
Co-Reward
& 16.7 & 13.3 & 47.0 & 75.2 & 31.8 & 73.7 \\
PC-Dr.DPO
& \textbf{20.0} & 13.3 & \textbf{53.0}
& \textbf{80.3} & 35.8 & \textbf{77.2} \\
\midrule

\multicolumn{7}{c}{\bfseries\itshape Qwen2.5-Math-1.5B-Instruct} \\
\addlinespace[0.5pt]
Base
& 11.1 & 8.0 & 46.8 & 73.2 & 25.8 & 55.2 \\
RENT
& 15.8 & 11.5 & 48.6 & 64.8 & 25.0 & 60.8 \\
TTRL
& 19.1 & \textbf{15.2} & 53.7
& 77.9 & 28.4 & 65.9 \\
Co-Reward
& 17.9 & 10.0 & 49.1 & 76.9 & 27.7 & \textbf{71.0} \\
PC-Dr.DPO
& \textbf{23.5} & 13.3 & \textbf{55.9}
& \textbf{78.2} & \textbf{29.5} & 69.2 \\
\midrule

\multicolumn{7}{c}{\bfseries\itshape Llama-3.2-3B-Instruct} \\
\addlinespace[0.5pt]
Base
& 7.4 & 0.3 & 18.5 & 40.3 & 26.4 & 43.5 \\
RENT
& $\ast$ & $\ast$ & $\ast$ & $\ast$ & 16.2 & 15.4 \\
TTRL
& 13.3 & $\ast$ & 29.9 & 57.4 & 27.9 & $\ast$ \\
Co-Reward
& 11.0 & $\ast$ & 28.9 & 49.3 & 26.8 & 35.8 \\
PC-Dr.DPO
& \textbf{16.7} & $\ast$ & \textbf{34.6}
& \textbf{58.4} & \textbf{28.0} & \textbf{45.8} \\
\bottomrule
\end{tabular*}
\endgroup

\vspace{-2mm}
\end{table}

\paragraph{Benchmarks.}
We compare PC-Dr.DPO with three TTRL baselines: the certainty-based method
RENT \cite{prabhudesai2025rent} and the ensemble-based methods TTRL and Co-Reward \cite{zuo2025ttrltesttimereinforcementlearning, zhang2025co}. Evaluation covers
four mathematical-reasoning benchmarks---AIME24, AIME25, AMC23, and
MATH500 \cite{hendrycks2021measuring}---and two
knowledge-intensive reasoning benchmarks, GPQA-Diamond \cite{rein2023gpqa} and MMLU-Pro \cite{wang2024mmlu}. Due to
computational constraints, we evaluate MMLU-Pro on its mathematics subset,
which contains 1,351 examples.

Test-time adaptation is performed on the full evaluation set, or on the
official test split when available. All methods use the same fixed
adaptation budget for each benchmark, and we report accuracy at the final checkpoint rather
than selecting the best intermediate checkpoint. This protocol avoids
favoring methods that improve early but subsequently collapse, reflecting
the practical setting in which the onset of degradation is not known in
advance.

\paragraph{Implementation details.}
We implement PC-Dr.DPO in the \textsc{veRL} framework~\cite{Sheng_2025}
and run all experiments on a single node with eight NVIDIA H100 GPUs.
We use Adam with a constant learning rate of $5\times10^{-7}$. For
PC-Dr.DPO, we sample $m=32$ rollouts per prompt and construct eight
preference pairs for each policy update. For TTRL and Co-Reward, we likewise
sample 32 rollouts per prompt and downsample them to eight training examples.
For RENT, we directly sample eight rollouts per prompt. This keeps the number
of examples entering each policy update consistent across methods. Rollouts
are generated with temperature $T=0.8$ and nucleus-sampling parameter
$p=0.9$. Additional hyperparameter settings are provided in
Appendix~B.1.

\begin{figure*}[!t]
    \centering
    \includegraphics[
        width=\textwidth,
        trim=10 8 10 8,
        clip
    ]{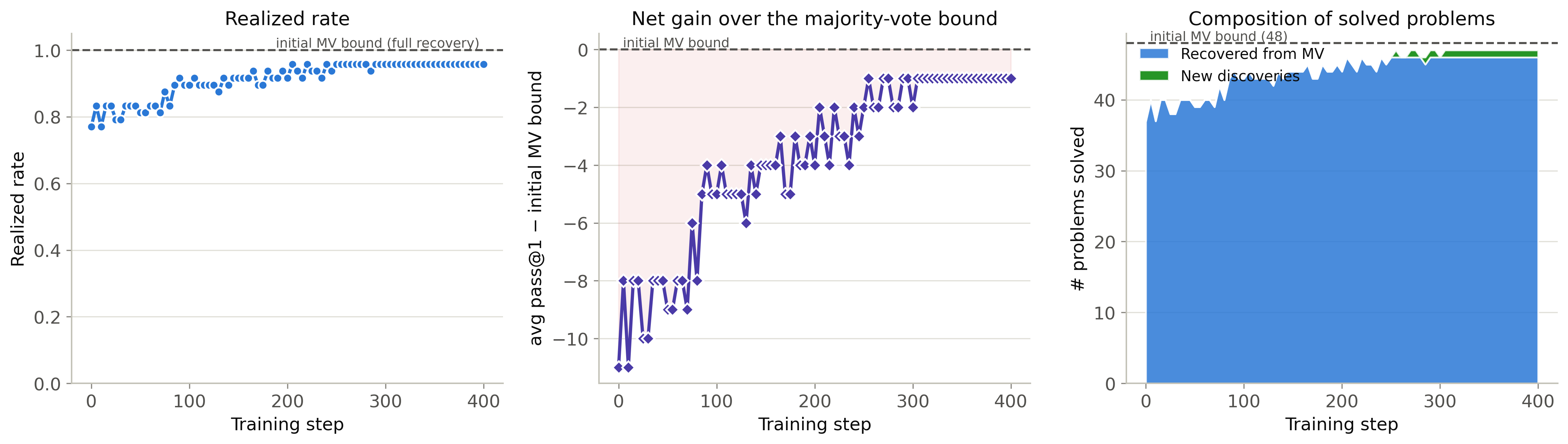}
    \caption{
    Adaptation dynamics of Qwen2.5-Math-1.5B-Instruct on the AMC benchmark.
    \textbf{Left:} fraction of problems in the initial majority-vote-solvable
    set that are recovered by the adapted policy.
    \textbf{Middle:} difference between the number of problems solved by the
    current policy and the initial majority-vote reference; the dashed line at
    zero denotes full recovery.
    \textbf{Right:} decomposition of the currently solved problems into those
    recovered from the initial majority-vote-solvable set and newly discovered
    solutions. The policy progressively recovers most initially available
    solutions while occasionally solving problems outside the initial set.
    }
    \label{fig:adaptation-dynamics}
    \vspace{-2mm}
\end{figure*}

\subsection{Main Results}
\label{sec:main_results}

Table~\ref{tab:main-results} reports test-time adaptation accuracy under
Avg@32 evaluation. Avg@32 denotes the mean correctness
over 32 sampled rollouts generated with temperature $T=0.8$ and
top-$p=0.9$. Results marked with $\ast$ indicate adaptation runs that
collapsed before the final checkpoint; representative failure cases are
provided in Appendix~C.2.

Across 24 of the 30 model--benchmark combinations, PC-Dr.DPO matches or
outperforms the strongest TTRL baseline. These results demonstrate the
benefit of prompt-conditional reweighting for reducing the optimization
effects of finite-rollout variation during test-time adaptation. 

\subsection{Ablation Studies}
\label{sec:ablation}

We conduct two ablation studies. First, we evaluate the contributions of
the reliability-conditioned temperature and score-based majority selection
using Qwen2.5-3B-Instruct on AMC and MATH500. Second, we study the
effect of the rollout-caching mechanism using Llama-3.2-3B-Instruct on
MATH500.

\paragraph{Temperature and majority selection.}
Table~\ref{tab:ablation-beta-majority} isolates the two principal design
choices in PC-Dr.DPO. The constant-temperature variant replaces the
reliability-conditioned schedule in Equation~\eqref{eq:beta-def} with
$\beta_x=\beta_{\mathrm{base}}$. The frequency-based variant replaces
score-based majority selection with standard answer-frequency voting while
retaining the adaptive temperature.

\begin{table}[b!]
\centering
\caption{Ablation of the reliability-conditioned temperature and
score-based majority selection on Qwen2.5-3B-Instruct. We report
Avg@32 accuracy (\%); the best adapted result is \textbf{bolded}.}
\label{tab:ablation-beta-majority}
\vspace{-1mm}
\small
\setlength{\tabcolsep}{4pt}
\renewcommand{\arraystretch}{0.92}
\begin{tabular}{@{}lcc@{}}
\toprule
Method & AMC & MATH500 \\
\midrule
Base model                     & 34.3 & 63.5 \\
PC-Dr.DPO w/ constant $\beta$  & 42.0 & 72.3 \\
PC-Dr.DPO w/ frequency MV      & 44.6 & 71.4 \\
PC-Dr.DPO                      & \textbf{45.8} & \textbf{73.3} \\
\bottomrule
\end{tabular}
\vspace{-2mm}
\end{table}

We further ablate the cache-replay mechanism using
Llama-3.2-3B-Instruct on MATH500. During adaptation, replaying previously
cached preference groups periodically anchors the policy to earlier
self-consistent solutions and mitigates self-supervised forgetting.
Removing this mechanism reduces Avg@32 accuracy from $58.4\%$ to $56.3\%$. Additional analysis is provided in
Appendix~C.3.

\subsection{Qualitative Analysis}
\label{sec:qualitative}

We analyze Qwen2.5-Math-1.5B-Instruct on AMC to better understand the
dynamics of test-time adaptation under majority-vote supervision. Both TTRL
and PC-Dr.DPO derive reinforcement signals from majority-voted responses
among the model's own rollouts. The set of problems solved by the initial
rollout majority therefore provides a useful reference for measuring how
effectively adaptation converts self-consistent solutions into improved
policy performance.

Empirically, we observe two distinct behaviors. For some problems,
adaptation successfully recovers solutions contained in the initial
majority-vote-solvable set and may additionally solve previously unsolved
problems, potentially through transfer of shared reasoning structure across
related prompts. For others, adaptation fails to recover the initially
available solutions and may even degrade performance. We attribute this
failure mode to the repeated reinforcement of incorrect population-majority
pseudo-labels, which can introduce systematic bias into the adapted policy.
Additional trajectories and representative examples on MATH500 across
multiple model backbones are provided in Appendix~C.4.

\section{Conclusion}
We analyzed majority-vote supervision in online-DPO TTRL, separating population-majority bias from prompt-dependent finite-rollout variation. To mitigate the latter, we introduced PC-Dr.DPO with reliability-conditioned within-prompt reweighting and cache replay. Across six benchmarks and five backbones, PC-Dr.DPO matches or exceeds the strongest TTRL baseline in 24 of 30 settings.

\bibliography{aaai2027}

\clearpage
\appendix
\section{Detailed Proofs}

\subsection{KL-DRO Derivation and Gradient Behavior}
\label{app:drdpo-proof}

Let $s=(x,y_w,y_l)$ denote a preference sample and recall
\begin{equation}
\label{eq:app-dpo-quantities}
\begin{aligned}
g_\theta(s)
&=
\log
\frac{\pi_\theta(y_w\mid x)}
     {\pi_{\mathrm{ref}}(y_w\mid x)}
-
\log
\frac{\pi_\theta(y_l\mid x)}
     {\pi_{\mathrm{ref}}(y_l\mid x)},\\
\nu_\theta(s)
&=
\sigma\!\left(\tau g_\theta(s)\right),
\qquad
h_\theta(s)
=
\log\nu_\theta(s).
\end{aligned}
\end{equation}
Let $p$ denote the empirical distribution over the preference pairs.

\paragraph{KL-constrained formulation.}
Consider the distributionally robust score
\begin{equation}
\label{eq:app-dro-primal}
\sup_{q:\,\mathrm{KL}(q\|p)\le\eta}
\mathbb E_{s\sim q}
\left[
h_\theta(s)
\right],
\end{equation}
where $\eta\ge0$ controls the allowable deviation from $p$. Assume
$q\ll p$ and let $r=dq/dp$ be the corresponding density ratio. Then
$r\ge0$, $\mathbb E_p[r]=1$, and
\[
\mathrm{KL}(q\|p)
=
\mathbb E_p[r\log r]
=
\mathbb E_p[\phi(r)],
\qquad
\phi(r):=r\log r-r+1.
\]
Hence, Equation~\eqref{eq:app-dro-primal} is equivalent to
\begin{equation}
\label{eq:app-dro-density}
\sup_{r\ge0}
\mathbb E_p[rh_\theta]
\quad
\text{s.t.}
\quad
\mathbb E_p[\phi(r)]\le\eta,
\qquad
\mathbb E_p[r]=1.
\end{equation}

Introduce a multiplier $\beta\ge0$ for the divergence constraint and
$\alpha\in\mathbb R$ for the normalization constraint. The Lagrangian is
\begin{align}
\mathcal L(r,\alpha,\beta)
&=
\mathbb E_p[rh_\theta]
-
\beta\bigl(\mathbb E_p[\phi(r)]-\eta\bigr)
+
\alpha\bigl(\mathbb E_p[r]-1\bigr)
\nonumber\\
&=
\beta\eta-\alpha
+
\beta\,
\mathbb E_p
\left[
r\frac{h_\theta+\alpha}{\beta}
-
\phi(r)
\right].
\label{eq:app-dro-lagrangian}
\end{align}

For $\beta>0$, the convex conjugate of
$\phi(r)=r\log r-r+1$ is
\[
\phi^*(u)
=
\sup_{r\ge0}
\{ru-\phi(r)\}
=
e^u-1,
\]
with maximizer $r^*(u)=e^u$. Maximizing
Equation~\eqref{eq:app-dro-lagrangian} over $r$ therefore gives
\begin{equation}
\label{eq:app-dual-alpha}
g(\alpha,\beta)
=
\beta\eta-\alpha
+
\beta\,
\mathbb E_p
\left[
e^{(h_\theta+\alpha)/\beta}-1
\right].
\end{equation}

Differentiating with respect to $\alpha$ yields
\[
-1+
\mathbb E_p
\left[
e^{(h_\theta+\alpha)/\beta}
\right]
=0,
\]
and thus
\begin{equation}
\label{eq:app-alpha-star}
\alpha^*
=
-\beta
\log
\mathbb E_p
\left[
e^{h_\theta/\beta}
\right].
\end{equation}
Substituting Equation~\eqref{eq:app-alpha-star} into
Equation~\eqref{eq:app-dual-alpha} gives
\begin{equation}
\label{eq:app-dro-dual}
\sup_{q:\,\mathrm{KL}(q\|p)\le\eta}
\mathbb E_q[h_\theta]
=
\inf_{\beta>0}
\left\{
\beta\eta
+
\beta
\log
\mathbb E_p
\left[
e^{h_\theta/\beta}
\right]
\right\}.
\end{equation}

For any $\beta>0$, the associated exponentially tilted distribution is
\begin{equation}
\label{eq:app-tilted-distribution}
q_{\theta,\beta}(s)
=
p(s)
\frac{
\exp\!\left(h_\theta(s)/\beta\right)
}{
\mathbb E_{s'\sim p}
\left[
\exp\!\left(h_\theta(s')/\beta\right)
\right]
}.
\end{equation}
Under the fixed-temperature approximation, we associate each prescribed
robustness radius $\eta$ with a temperature $\beta^\star(\eta)$, which is
treated as independent of $\theta$ and held fixed during the policy update.

\paragraph{Fixed-temperature objective.}
Assume that $\beta^\star(\eta)$ minimizes the dual objective in
Equation~\eqref{eq:app-dro-dual} under this fixed-temperature
parameterization. Then
\begin{equation}
\label{eq:app-dro-optimal}
\sup_{q:\,\mathrm{KL}(q\|p)\le\eta}
\mathbb E_q[h_\theta]
=
\beta^\star(\eta)\eta
+
\beta^\star(\eta)
\log
\mathbb E_{s\sim p}
\left[
\exp\!\left(
\frac{h_\theta(s)}{\beta^\star(\eta)}
\right)
\right].
\end{equation}
Because $\eta$ and $\beta^\star(\eta)$ are fixed during the policy update,
the term $\beta^\star(\eta)\eta$ is independent of $\theta$ and can be
omitted. Writing $\beta=\beta^\star(\eta)$ and negating the remaining
robust score gives the Dr.DPO loss
\begin{equation}
\label{eq:app-dro-loss}
\mathcal L_{\mathrm{DrDPO}}(\theta)
=
-\beta
\log
\mathbb E_{s\sim p}
\left[
\exp\!\left(
\frac{h_\theta(s)}{\beta}
\right)
\right].
\end{equation}

\paragraph{Gradient.}
Define
\begin{equation}
\label{eq:app-dro-weight}
w_\beta(s)
:=
\frac{
\exp\!\left(h_\theta(s)/\beta\right)
}{
\mathbb E_{s'\sim p}
\left[
\exp\!\left(h_\theta(s')/\beta\right)
\right]
}.
\end{equation}
Since
\begin{equation}
\label{eq:app-score-gradient}
\nabla_\theta h_\theta(s)
=
\tau
\bigl(1-\nu_\theta(s)\bigr)
\nabla_\theta g_\theta(s),
\end{equation}
differentiating Equation~\eqref{eq:app-dro-loss} gives
\begin{equation}
\label{eq:app-dro-gradient}
\nabla_\theta
\mathcal L_{\mathrm{DrDPO}}(\theta)
=
-\tau\,
\mathbb E_{s\sim p}
\left[
w_\beta(s)
\bigl(1-\nu_\theta(s)\bigr)
\nabla_\theta g_\theta(s)
\right].
\end{equation}

\paragraph{Effect of the temperature.}
Consider a finite collection of preference pairs
$\mathcal S=\{s_i\}_{i=1}^{N}$ and
let
\[
\nu_i:=\nu_\theta(s_i).
\]
The normalized exponentially tilted weights are
\begin{equation}
\label{eq:app-temperature-weights}
\overline w_i(\beta)
=
\frac{\nu_i^{1/\beta}}
     {\sum_{k=1}^{N}\nu_k^{1/\beta}}.
\end{equation}
The corresponding gradient is
\begin{equation}
\label{eq:app-temperature-gradient}
\nabla_\theta\mathcal L(\theta)
=
-\tau
\sum_{i=1}^{N}
\overline w_i(\beta)
\bigl(1-\nu_i\bigr)
\nabla_\theta g_\theta(s_i).
\end{equation}

Define the weighted residual logistic factor
\begin{equation}
\label{eq:app-aggregate-factor}
A(\beta)
:=
\sum_{i=1}^{N}
\overline w_i(\beta)
\bigl(1-\nu_i\bigr)
=
1-
\sum_{i=1}^{N}
\overline w_i(\beta)\nu_i.
\end{equation}
Let $t=1/\beta$. Then
\[
\overline w_i(t)
=
\frac{\exp(t\log\nu_i)}
     {\sum_{k=1}^{N}\exp(t\log\nu_k)}.
\]
Differentiating the weighted mean of $\nu_i$ gives
\begin{equation}
\label{eq:app-temperature-derivative}
\frac{d}{dt}
\sum_{i=1}^{N}
\overline w_i(t)\nu_i
=
\operatorname{Cov}_{\overline w(t)}
\left(
\nu_i,\log\nu_i
\right)
\ge 0,
\end{equation}
where the inequality follows because $\log\nu$ is increasing in $\nu$.
Consequently, $A(\beta)$ is nondecreasing in $\beta$, or equivalently,
decreases as $\beta$ decreases.

If the pairwise margin gradients satisfy
\[
\left\|
\nabla_\theta g_\theta(s_i)
\right\|
\le G
\qquad
\text{for all }i,
\]
then
\begin{equation}
\label{eq:app-gradient-bound}
\left\|
\nabla_\theta\mathcal L(\theta)
\right\|
\le
\tau G\,A(\beta).
\end{equation}
Thus, decreasing $\beta$ decreases an upper bound on the update magnitude.

\subsection{Proof of Theorem~1}
\label{app:proof-generalization}

We first restate the uniform empirical-Bernstein result used in the proof.

\begin{restatedlemma}[Uniform empirical-Bernstein bound; adapted from
\cite{menon2021statistical}]
\label{lem:menon}
Let $\ell:\{0,1\}\times\mathbb R\to[0,1]$, let
$\mathcal F$ be a class of predictors $f:\mathcal S\to\mathbb R$, and let
$a:\mathcal S\to\Delta_2$ be a fixed measurable map. Define
\[
\ell(f(s))
:=
\begin{pmatrix}
\ell(1,f(s))\\
\ell(0,f(s))
\end{pmatrix}
\]
and the induced loss class
\[
\mathcal H_a
:=
\left\{
h_f^a:\mathcal S\to[0,1]
\ \middle|\
h_f^a(s)=a(s)^\top\ell(f(s)),
\ f\in\mathcal F
\right\}.
\]
Let
\[
\mathcal M_N(a)
:=
\mathcal N_\infty
\left(
\frac{1}{N},
\mathcal H_a,
2N
\right),
\]
where $\mathcal N_\infty(\epsilon,\mathcal H,m)$ denotes the uniform
$\ell_\infty$ covering number of $\mathcal H$ on $m$ points.

For a sample $S=\{s_n\}_{n=1}^{N}$, define
\begin{equation}
\label{eq:app-empirical-risk-a}
\widehat R_a(f;S)
:=
\frac{1}{N}
\sum_{n=1}^{N}
a(s_n)^\top\ell(f(s_n)),
\end{equation}
and let
\begin{equation}
\label{eq:app-variance-a}
\mathbb V_{N,a}(f;S)
:=
\frac{1}{N}
\sum_{n=1}^{N}
\left(
a(s_n)^\top\ell(f(s_n))
-
\widehat R_a(f;S)
\right)^2.
\end{equation}
Then, for any $\delta\in(0,1)$, with probability at least $1-\delta$
over $S\sim\mathcal P^N$, every $f\in\mathcal F$ satisfies
\begin{equation}
\label{eq:app-menon-bound}
\begin{aligned}
\mathbb E_{s\sim\mathcal P}
\left[
a(s)^\top\ell(f(s))
\right]
\le\;&
\widehat R_a(f;S)\\
&+
O\left(
\sqrt{
\frac{
\mathbb V_{N,a}(f;S)
\log(\mathcal M_N(a)/\delta)
}{N}
}
\right.\\[-1mm]
&\left.\hspace{8mm}
+
\frac{\log(\mathcal M_N(a)/\delta)}{N}
\right).
\end{aligned}
\end{equation}
\end{restatedlemma}

\begin{proof}[Proof of Theorem~1]
For a preference sample $s=(x,y_w,y_l)$, write
\[
\ell(f(s))
:=
\begin{pmatrix}
\ell(1,f(s))\\
\ell(0,f(s))
\end{pmatrix}
\in[0,1]^2.
\]
Let $p^*(s),\hat p(s)\in\Delta_2$ denote the true and pseudo-preference
distributions associated with $s$, respectively. 

Define the true population risk, pseudo-preference population risk, and
pseudo-preference empirical risk by
\begin{align}
R(f)
&:=
\mathbb E_{s\sim\mathcal P}
\left[
p^*(s)^\top\ell(f(s))
\right],
\label{eq:app-true-risk}\\
\widetilde R(f)
&:=
\mathbb E_{s\sim\mathcal P}
\left[
\hat p(s)^\top\ell(f(s))
\right],
\label{eq:app-pseudo-pop-risk}\\
\widetilde R(f;S)
&:=
\frac{1}{N}
\sum_{n=1}^{N}
\hat p(s_n)^\top\ell(f(s_n)).
\label{eq:app-pseudo-emp-risk}
\end{align}

Consider the induced pseudo-preference loss class
\[
\widetilde{\mathcal H}
:=
\left\{
\widetilde h_f:\mathcal S\to[0,1]
\ \middle|\
\widetilde h_f(s)
=
\hat p(s)^\top\ell(f(s)),
\ f\in\mathcal F
\right\},
\]
and define
\[
\widetilde{\mathcal M}_N
:=
\mathcal N_\infty
\left(
\frac{1}{N},
\widetilde{\mathcal H},
2N
\right).
\]
Because $\hat p(s)\in\Delta_2$ and
$\ell(f(s))\in[0,1]^2$, every function in
$\widetilde{\mathcal H}$ takes values in $[0,1]$.

Let $\widetilde{\mathbb V}_N(f;S)$ denote the empirical variance of
\[
\left\{
\hat p(s_n)^\top\ell(f(s_n))
\right\}_{n=1}^{N}.
\]
Applying the empirical-Bernstein bound with $a(s)=\hat p(s)$ gives that, with
probability at least $1-\delta$ over $S\sim\mathcal P^N$, every
$f\in\mathcal F$ satisfies
\begin{equation}
\label{eq:app-pseudo-generalization}
\begin{aligned}
\widetilde R(f)
\le\;&
\widetilde R(f;S)\\
&+
O\left(
\sqrt{
\frac{
\widetilde{\mathbb V}_N(f;S)
\log(\widetilde{\mathcal M}_N/\delta)
}{N}
}
\right.\\[-1mm]
&\left.\hspace{8mm}
+
\frac{\log(\widetilde{\mathcal M}_N/\delta)}{N}
\right).
\end{aligned}
\end{equation}

It remains to control the discrepancy between the true and
pseudo-preference population risks. Let
\[
d(s)
:=
\hat p(s)-p^*(s)
=
\begin{pmatrix}
d_1(s)\\
d_0(s)
\end{pmatrix}.
\]
Since both $\hat p(s)$ and $p^*(s)$ are probability vectors,
\[
d_1(s)+d_0(s)=0.
\]
Therefore,
\begin{align}
\left|
d(s)^\top\ell(f(s))
\right|
&=
\left|
d_1(s)\ell(1,f(s))
+
d_0(s)\ell(0,f(s))
\right|
\nonumber\\
&=
\left|
d_1(s)
\left(
\ell(1,f(s))-\ell(0,f(s))
\right)
\right|
\nonumber\\
&\le
|d_1(s)|
\nonumber\\
&=
\frac{1}{2}
\left(
|d_1(s)|+|d_0(s)|
\right)
\nonumber\\
&=
\frac{1}{2}
\left\|
\hat p(s)-p^*(s)
\right\|_1.
\label{eq:app-pointwise-tv}
\end{align}
The inequality follows from
$\ell(1,f(s)),\ell(0,f(s))\in[0,1]$, which implies
\[
\left|
\ell(1,f(s))-\ell(0,f(s))
\right|
\le 1.
\]

Taking expectations in Equation~\eqref{eq:app-pointwise-tv} yields
\begin{align}
\left|
\widetilde R(f)-R(f)
\right|
&=
\left|
\mathbb E_{s\sim\mathcal P}
\left[
\bigl(\hat p(s)-p^*(s)\bigr)^\top
\ell(f(s))
\right]
\right|
\nonumber\\
&\le
\mathbb E_{s\sim\mathcal P}
\left[
\left|
\bigl(\hat p(s)-p^*(s)\bigr)^\top
\ell(f(s))
\right|
\right]
\nonumber\\
&\le
\frac{1}{2}
\mathbb E_{s\sim\mathcal P}
\left[
\left\|
\hat p(s)-p^*(s)
\right\|_1
\right].
\label{eq:app-risk-tv}
\end{align}
Consequently,
\begin{equation}
\label{eq:app-risk-decomposition}
R(f)
\le
\widetilde R(f)
+
\frac{1}{2}
\mathbb E_{s\sim\mathcal P}
\left[
\left\|
\hat p(s)-p^*(s)
\right\|_1
\right].
\end{equation}
\end{proof}

\subsection{Total-Variation Identity for Deterministic Binary Preferences}
\label{app:tv-majority-identity}
\begin{appendixlemma}[Total-variation identity for deterministic binary preferences]
\label{lem:tv-majority-error}
Assume both
the true and pseudo-preference distributions are deterministic. Then
\begin{equation}
\label{eq:app-tv-indicator}
\frac{1}{2}
\left\|
\hat p(s)-p^*(s)
\right\|_1
=
\mathbf 1
\left\{
\widehat a_{m,b}(x)\neq a^\star(x)
\right\}.
\end{equation}
Consequently,
\begin{equation}
\label{eq:app-tv-majority-expectation}
\frac{1}{2}
\mathbb E_{s\sim\mathcal P}
\left[
\left\|
\hat p(s)-p^*(s)
\right\|_1
\right]
=
\mathbb E_x
\left[
U_{m,b}(x)
\right].
\end{equation}
\end{appendixlemma}

\begin{proof}
Let $z^*(s)\in\{0,1\}$ denote the true preference label and let
$\widehat z(s)\in\{0,1\}$ denote the pseudo-preference label induced by the
sampled-majority answer. Under the deterministic binary construction,
\[
p^*(s)=e_{z^*(s)},
\qquad
\hat p(s)=e_{\widehat z(s)},
\]
where $e_0=(1,0)^\top$ and $e_1=(0,1)^\top$. Hence,
\begin{align}
\frac{1}{2}
\left\|
\hat p(s)-p^*(s)
\right\|_1
&=
\frac{1}{2}
\left\|
e_{\widehat z(s)}-e_{z^*(s)}
\right\|_1
\nonumber\\
&=
\mathbf 1
\left\{
\widehat z(s)\neq z^*(s)
\right\}
\nonumber\\
&=
\mathbf 1
\left\{
\widehat a_{m,b}(x)\neq a^\star(x)
\right\}.
\label{eq:app-tv-one-hot}
\end{align}

Taking expectations and applying the law of total expectation gives
\begin{align}
\frac{1}{2}
\mathbb E_{s\sim\mathcal P}
\left[
\left\|
\hat p(s)-p^*(s)
\right\|_1
\right]
&=
\mathbb E_x
\left[
\Pr\!\left(
\widehat a_{m,b}(x)\neq a^\star(x)
\mid x
\right)
\right]
\nonumber\\
&=
\mathbb E_x
\left[
U_{m,b}(x)
\right].
\end{align}
\end{proof}

\subsection{Proof of Majority-Vote Concentration}
\label{app:proof-majority-concentration}

\begin{proof}
Fix a prompt $x$ and let
\[
Z_i
:=
\mathbf 1
\left\{
\operatorname{ans}(y_i)=a^\star(x)
\right\},
\qquad
y_i\overset{\mathrm{i.i.d.}}{\sim}\pi_b(\cdot\mid x),
\]
for $i=1,\ldots,m$. Conditional on $x$, the variables
$Z_1,\ldots,Z_m$ are i.i.d.\ Bernoulli random variables satisfying
\[
\mathbb E[Z_i\mid x]
=
\rho_b(x).
\]
Define their empirical mean by
\[
\overline Z_m
:=
\frac{1}{m}\sum_{i=1}^{m}Z_i.
\]

Under the binary-answer abstraction, if $\rho_b(x)>1/2$, then the
population-majority answer is $a^\star(x)$. Therefore, disagreement between
the sampled and population majorities implies
$\overline Z_m\le 1/2$. Hence, Hoeffding's inequality gives
\begin{align}
V_{m,b}(x)
&\le
\Pr\!\left(
\overline Z_m\le\frac{1}{2}
\mid x
\right)
\nonumber\\
&=
\Pr\!\left(
\overline Z_m-\rho_b(x)
\le
-\left(\rho_b(x)-\frac{1}{2}\right)
\middle| x
\right)
\nonumber\\
&\le
\exp\!\left(
-2m\left(\rho_b(x)-\frac{1}{2}\right)^2
\right).
\label{eq:app-concentration-correct-mode}
\end{align}

If instead $\rho_b(x)<1/2$, then the population-majority answer is
incorrect. Disagreement with that population majority implies
$\overline Z_m\ge 1/2$, and Hoeffding's inequality yields
\begin{align}
V_{m,b}(x)
&\le
\Pr\!\left(
\overline Z_m\ge\frac{1}{2}
\mid x
\right)
\nonumber\\
&=
\Pr\!\left(
\overline Z_m-\rho_b(x)
\ge
\frac{1}{2}-\rho_b(x)
\middle| x
\right)
\nonumber\\
&\le
\exp\!\left(
-2m\left(\frac{1}{2}-\rho_b(x)\right)^2
\right).
\label{eq:app-concentration-incorrect-mode}
\end{align}
For even $m$, these inclusions remain valid under any deterministic
tie-breaking rule. Combining the two cases gives
\[
V_{m,b}(x)
\le
\exp\!\left(
-2m\left|\rho_b(x)-\frac{1}{2}\right|^2
\right).
\]

For completeness, under the binary abstraction,
\[
U_{m,b}(x)
=
B_b(x)
+
\bigl(1-2B_b(x)\bigr)V_{m,b}(x).
\]
Thus, when $\rho_b(x)\ge 1/2+\gamma$, we have $B_b(x)=0$ and
\[
U_{m,b}(x)
=
V_{m,b}(x)
\le
e^{-2m\gamma^2}.
\]
When $\rho_b(x)\le 1/2-\gamma$, we have $B_b(x)=1$ and
\[
U_{m,b}(x)
=
1-V_{m,b}(x)
\ge
1-e^{-2m\gamma^2}.
\]
\end{proof}

\subsection{Proof of Bias--Variation Separation}
\label{app:proof-bias-variation-corollary}

\begin{proof}
Under the binary-answer abstraction, the exact majority-error decomposition
gives
\begin{equation}
\label{eq:app-bias-variation-decomposition}
U_{m,b}(x)-B_b(x)
=
\bigl(1-2B_b(x)\bigr)V_{m,b}(x),
\end{equation}
where
\[
B_b(x)
=
\mathbf 1
\left\{
\rho_b(x)<\frac12
\right\}.
\]
Since $B_b(x)\in\{0,1\}$, we have
$\lvert 1-2B_b(x)\rvert=1$, and therefore
\begin{equation}
\label{eq:app-error-bias-difference}
\left|
U_{m,b}(x)-B_b(x)
\right|
=
V_{m,b}(x).
\end{equation}

Define
\[
\alpha_b
:=
\mathbb E_x[B_b(x)]
=
\Pr_x\!\left(
\rho_b(x)<\frac12
\right).
\]
Under the uniform gap assumption
\[
\left|
\rho_b(x)-\frac12
\right|
\ge
\gamma
\qquad
\text{almost surely},
\]
the majority-vote concentration result proved above yields
\[
V_{m,b}(x)
\le
e^{-2m\gamma^2}
\qquad
\text{almost surely}.
\]
Hence,
\begin{align}
\left|
\mathbb E_x[U_{m,b}(x)]-\alpha_b
\right|
&=
\left|
\mathbb E_x
\left[
U_{m,b}(x)-B_b(x)
\right]
\right|
\nonumber\\
&\le
\mathbb E_x
\left[
\left|
U_{m,b}(x)-B_b(x)
\right|
\right]
\nonumber\\
&=
\mathbb E_x[V_{m,b}(x)]
\nonumber\\
&\le
e^{-2m\gamma^2}.
\label{eq:app-error-to-bias}
\end{align}
In particular,
\begin{equation}
\label{eq:app-majority-error-upper}
\mathbb E_x[U_{m,b}(x)]
\le
\alpha_b+e^{-2m\gamma^2}.
\end{equation}

The majority-vote generalization bound gives, with probability at least
$1-\delta$ over $S\sim\mathcal P^N$, simultaneously for every
$f\in\mathcal F$,
\[
R(f)
\le
\widetilde R(f;S)
+
\mathbb E_x[U_{m,b}(x)]
+
\mathcal G_N(f,\delta;S).
\]
Substituting Equation~\eqref{eq:app-majority-error-upper} yields
\[
R(f)
\le
\widetilde R(f;S)
+
\alpha_b
+
e^{-2m\gamma^2}
+
\mathcal G_N(f,\delta;S),
\]
which proves the result.
\end{proof}
\section{Implementation Details}
\subsection{Hyperparameters}
\label{app:hyperparameters}

\begin{table}
\centering
\caption{Hyperparameters used in our experiments.}
\label{tab:hyperparameters}
\scriptsize
\setlength{\tabcolsep}{3.5pt}
\renewcommand{\arraystretch}{1.05}
\begin{tabularx}{\columnwidth}{
|>{\raggedright\arraybackslash}p{0.40\columnwidth}
|>{\raggedright\arraybackslash}X|}
\hline
\textbf{Hyperparameter} & \textbf{Value} \\
\hline

Max prompt length
& 1,024 \\
\hline

Max response length
& 3,072 \\
\hline

Training batch size
&
\begin{tabular}[t]{@{}l@{}}
24 AIME24 \\
24 AIME25 \\
16 AMC \\
32 MATH500 \\
16 GPQA-Diamond \\
32 MMLU-Pro Math
\end{tabular}
\\
\hline

Dataset size
&
\begin{tabular}[t]{@{}l@{}}
30 AIME24 \\
30 AIME25 \\
83 AMC \\
500 MATH500 \\
198 GPQA-Diamond \\
1,351 MMLU-Pro Math
\end{tabular}
\\
\hline

Total epochs
&
\begin{tabular}[t]{@{}l@{}}
240 AIME24 \\
240 AIME25 \\
80 AMC \\
40 MATH500 \\
40 GPQA-Diamond \\
80 MMLU-Pro Math
\end{tabular}
\\
\hline

PPO mini-batch size
& 4 \\
\hline

PPO micro-batch size
& 8 \\
\hline

Optimizer
& Adam \\
\hline

Learning rate
& $5\times10^{-7}$ \\
\hline

DPO inverse temperature $\tau$
& 0.1 \\
\hline

Base DRO temperature $\beta_{\mathrm{base}}$
& 2 \\
\hline

Preference pairs per prompt
& 8 \\
\hline

Rollout scoring strategy
& \texttt{implicit\_dpo\_reward} \\
\hline

Pair-construction strategy
& \texttt{random} \\
\hline

Training rollouts per prompt
& 32 \\
\hline

Training sampling
&
\begin{tabular}[t]{@{}l@{}}
Temperature: 0.8 \\
Top-$p$: 0.9 \\
Top-$k$: $-1$
\end{tabular}
\\
\hline

Validation rollouts per prompt
& 32 \\
\hline

Validation sampling
&
\begin{tabular}[t]{@{}l@{}}
Temperature: 0.8 \\
Top-$p$: 0.9 \\
Top-$k$: $-1$
\end{tabular}
\\
\hline

\end{tabularx}
\end{table}

For PC-Dr.DPO, \texttt{ppo\_micro\_batch\_size} is set equal to the number
of preference pairs constructed per prompt, ensuring that each prompt group
remains intact during within-prompt robust reweighting.

\section{More Results} 
\subsection{Effect of the Rollout Count}
\label{app:rollout-count}

\begin{figure*}[t]
    \centering

    \begin{minipage}[t]{0.495\textwidth}
        \centering
        \textbf{(a) Pseudo-label Accuracy}\par
    \vspace{1.5mm}
        \includegraphics[
            width=\linewidth,
            trim=0 0 0 0,
            clip
        ]{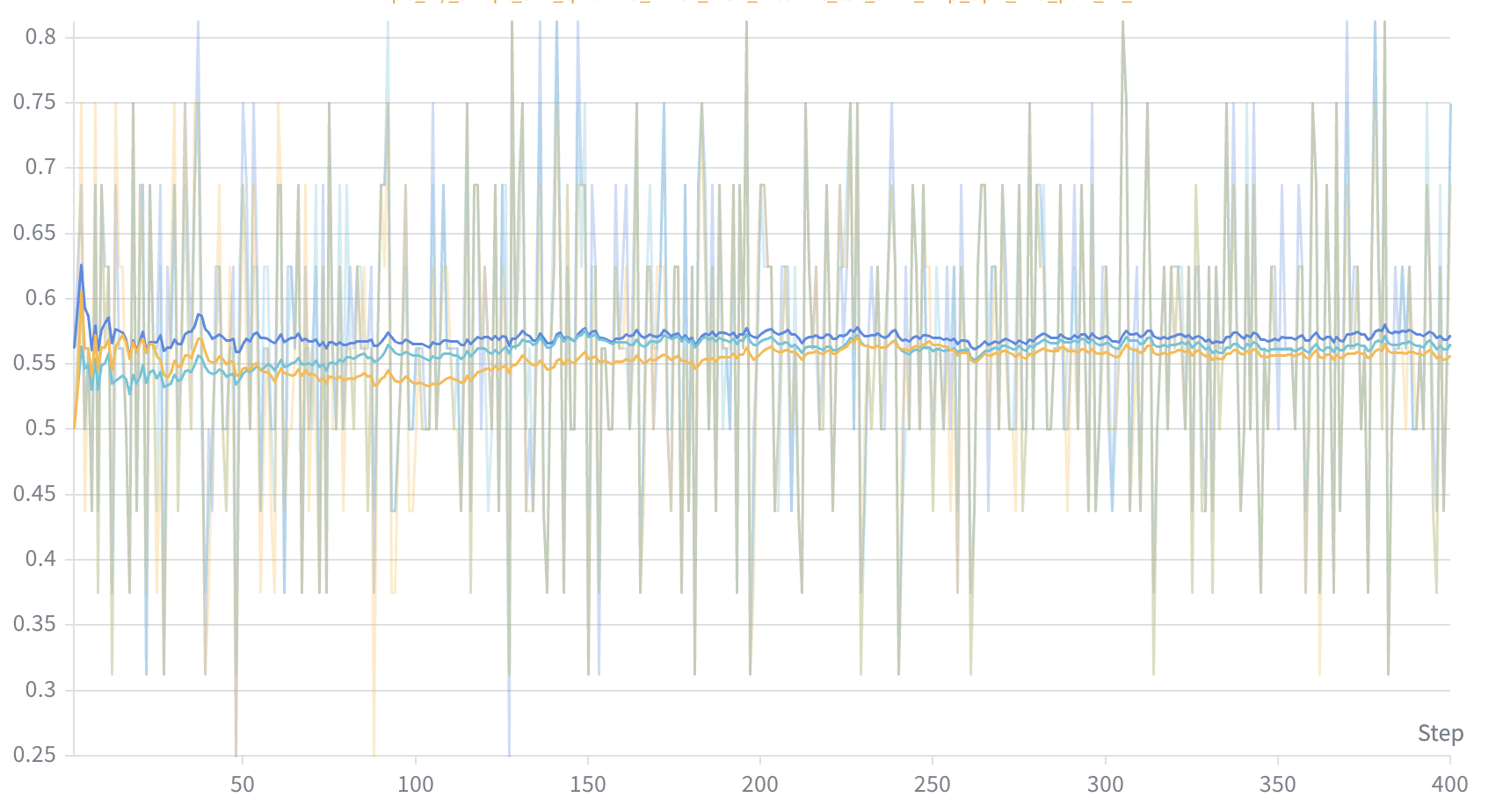}
    \end{minipage}
    \hfill
    \begin{minipage}[t]{0.495\textwidth}
        \centering
        \textbf{(b) Validation Avg@32}\par
    \vspace{1.5mm}
        \includegraphics[
            width=\linewidth,
            trim=0 0 0 0,
            clip
        ]{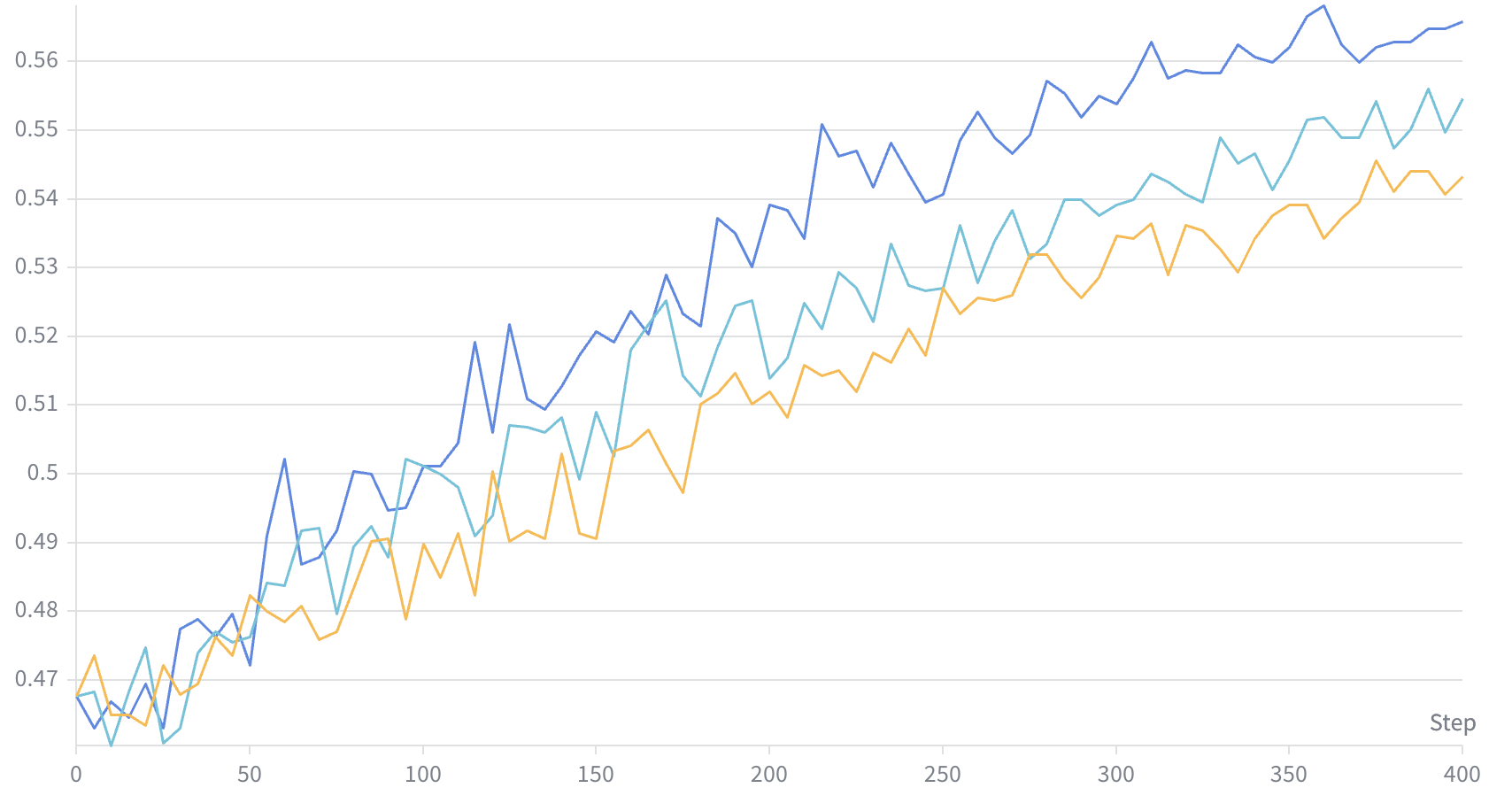}
    \end{minipage}

    \vspace{1mm}

    {\small
    \textcolor{mdeepblue}{\rule{10mm}{1.5pt}}\;$m=64$
    \qquad
    \textcolor{mblue}{\rule{10mm}{1.5pt}}\;$m=32$
    \qquad
    \textcolor{myellow}{\rule{10mm}{1.5pt}}\;$m=16$
    }

    \caption{
    Effect of the rollout count $m$ during test-time adaptation of
    Qwen2.5-Math-1.5B-Instruct on AMC.
    \textbf{Left:} pseudo-label accuracy across adaptation steps. The faint
    trajectories show the per-step measurements, while the darker curves
    show their smoothed trends.
    \textbf{Right:} validation Avg@32 accuracy throughout adaptation.
    Increasing $m$ reduces the variability of sampled pseudo-labels and
    produces more stable downstream improvement.
    }
    \label{fig:rollout-count-analysis}
\end{figure*}

\begin{figure*}[t]
    \centering

    \begin{minipage}[t]{0.325\textwidth}
        \centering
        \textbf{(a) Pseudo-label Accuracy}\par
    \vspace{1.5mm}
        \includegraphics[
            width=\linewidth,
            trim=0 0 0 0,
            clip
        ]{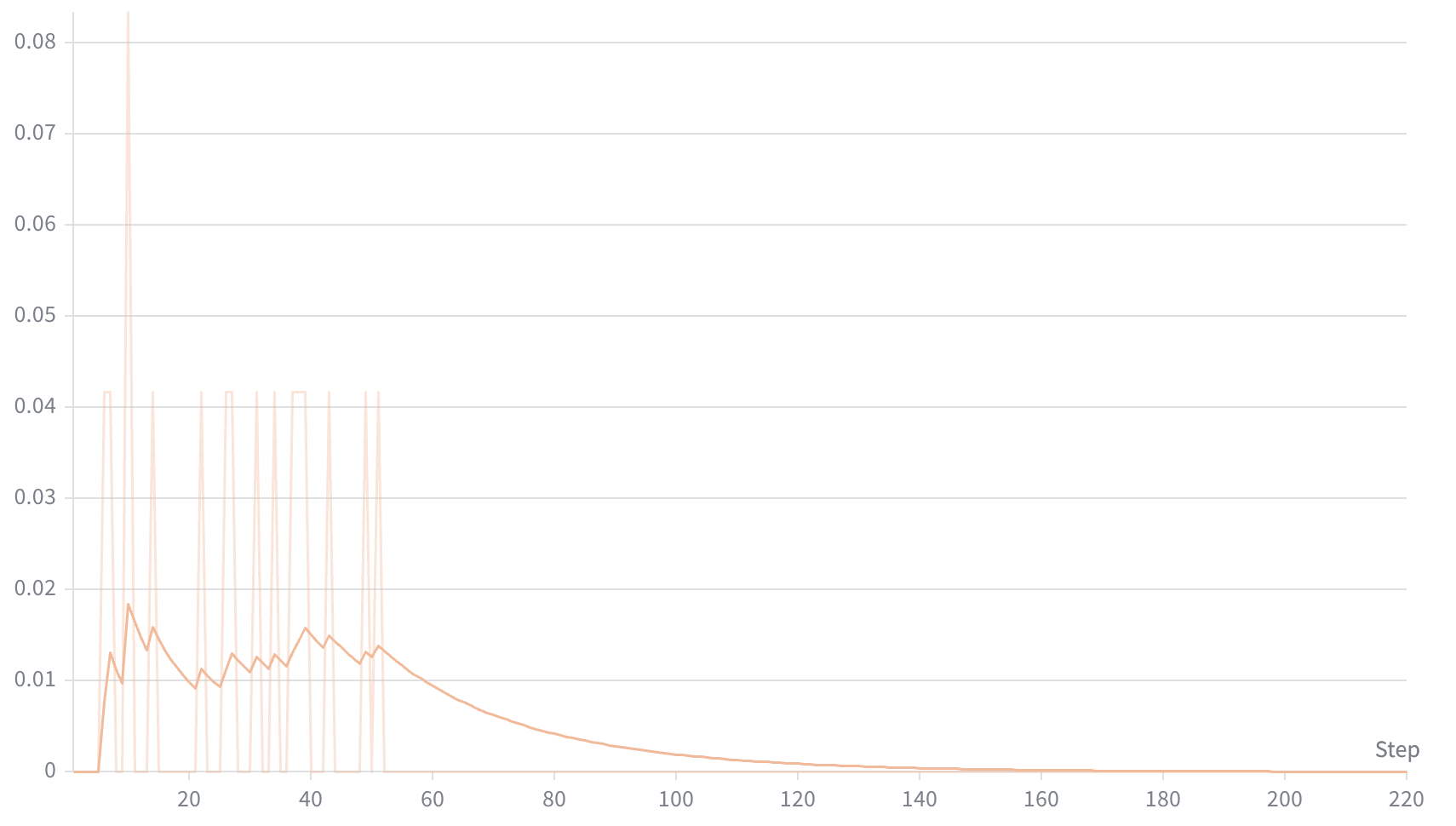}
    \end{minipage}
    \hfill
    \begin{minipage}[t]{0.325\textwidth}
        \centering
        \textbf{(b) Validation Avg@32}\par
    \vspace{1.5mm}
        \includegraphics[
            width=\linewidth,
            trim=0 0 0 0,
            clip
        ]{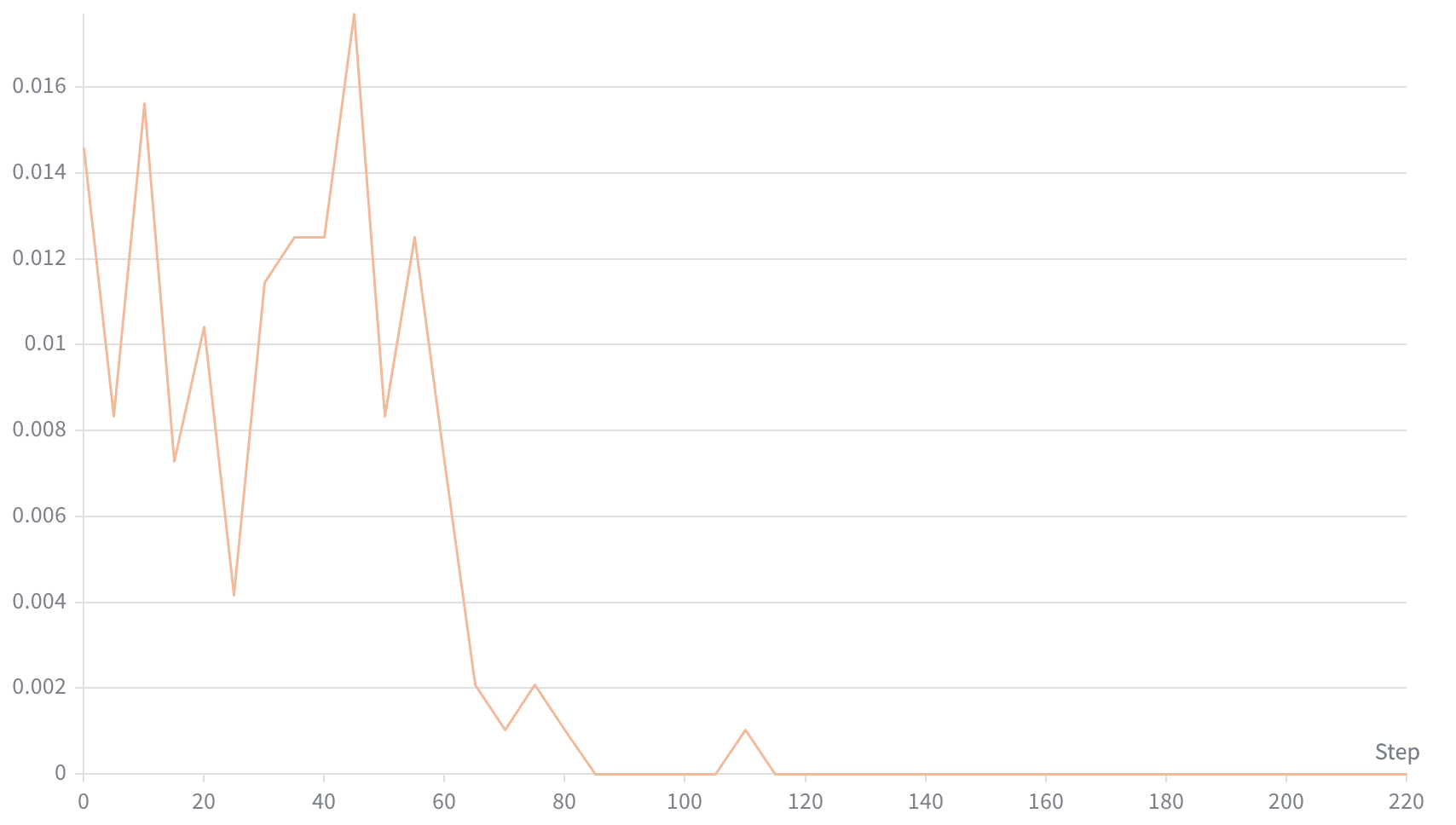}
    \end{minipage}
    \hfill
    \begin{minipage}[t]{0.325\textwidth}
        \centering
        \textbf{(c) Modal-Answer Fraction}\par
    \vspace{1.5mm}
        \includegraphics[
            width=\linewidth,
            trim=0 0 0 0,
            clip
        ]{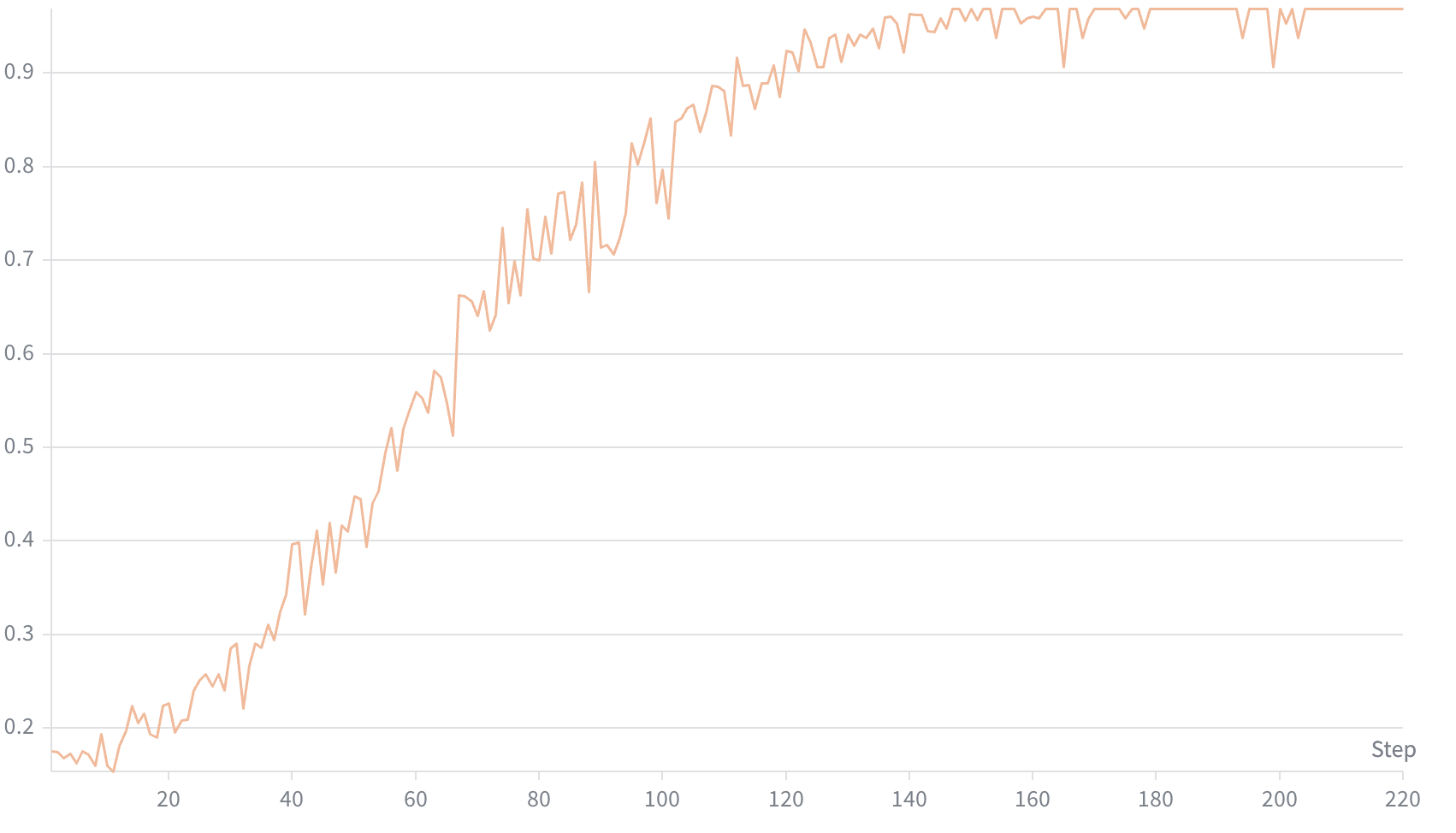}
    \end{minipage}

    \caption{
    Self-reinforcing collapse of Qwen2.5-1.5B-Instruct during test-time
    adaptation on AIME25.
    \textbf{Left:} accuracy of the sampled pseudo-labels relative to the
    ground-truth answers.
    \textbf{Middle:} validation Avg@32 accuracy, which eventually decreases
    to zero.
    \textbf{Right:} average modal-answer fraction within each rollout group.
    Although the modal-answer fraction steadily approaches one, the
    pseudo-label accuracy remains close to zero. Thus, the policy becomes
    increasingly confident in predominantly incorrect pseudo-labels,
    reinforcing systematic population-majority bias and ultimately producing
    adaptation collapse.
    }
    \label{fig:aime25-collapse}
\end{figure*}

Figure~\ref{fig:rollout-count-analysis} compares rollout counts
$m\in\{16,32,64\}$ for Qwen2.5-Math-1.5B-Instruct on AMC. Larger rollout
counts reduce variation in the sampled pseudo-labels and yield more stable
validation trajectories. This observation is consistent with our analysis:
increasing $m$ suppresses finite-rollout variation.

\subsection{Self-Reinforcing Collapse under Incorrect Population Majorities}
\label{app:self-reinforcing-collapse}
Figure~\ref{fig:aime25-collapse} shows the adaptation dynamics of
Qwen2.5-1.5B-Instruct on AIME25. The sampled pseudo-label accuracy remains
close to zero, indicating that most prompt-level supervision signals are
incorrect. Nevertheless, the modal-answer fraction increases throughout
training and eventually approaches one. The policy therefore becomes more
internally consistent without becoming more correct. Repeated optimization
against these increasingly decisive but incorrect pseudo-labels reinforces
population-majority bias, while validation Avg@32 accuracy ultimately
collapses to zero.

\subsection{Analysis of Cache Replay}
\label{app:cache-replay}

\begin{figure*}[t]
    \centering

    \begin{minipage}[t]{0.40\textwidth}
        \centering
        \textbf{(a) Cache-Replay Dynamics}\par
    \vspace{1.5mm}
        \includegraphics[
            width=\linewidth,
            trim=0 0 0 0,
            clip
        ]{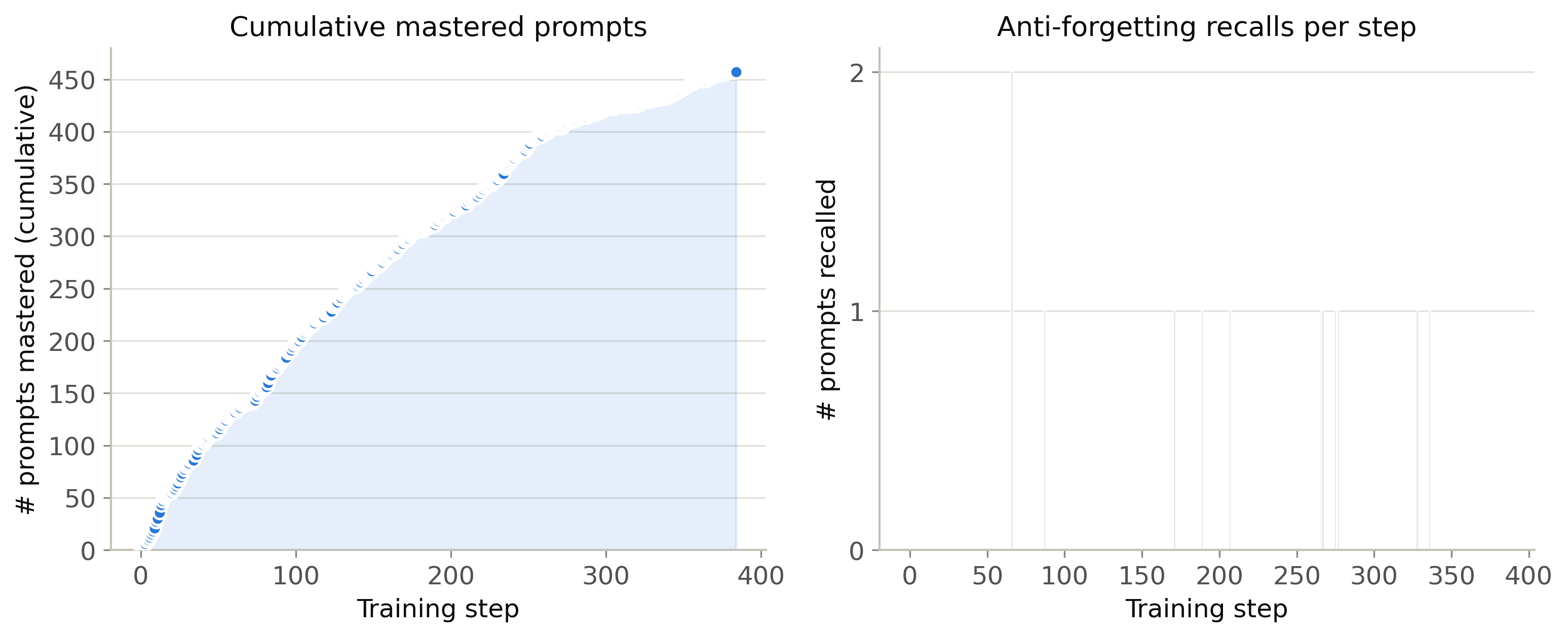}
    \end{minipage}
    \hfill
    \begin{minipage}[t]{0.285\textwidth}
        \centering
        \textbf{(b) Pseudo-label Accuracy}\par
    \vspace{1.5mm}
        \includegraphics[
            width=\linewidth,
            trim=0 0 0 0,
            clip
        ]{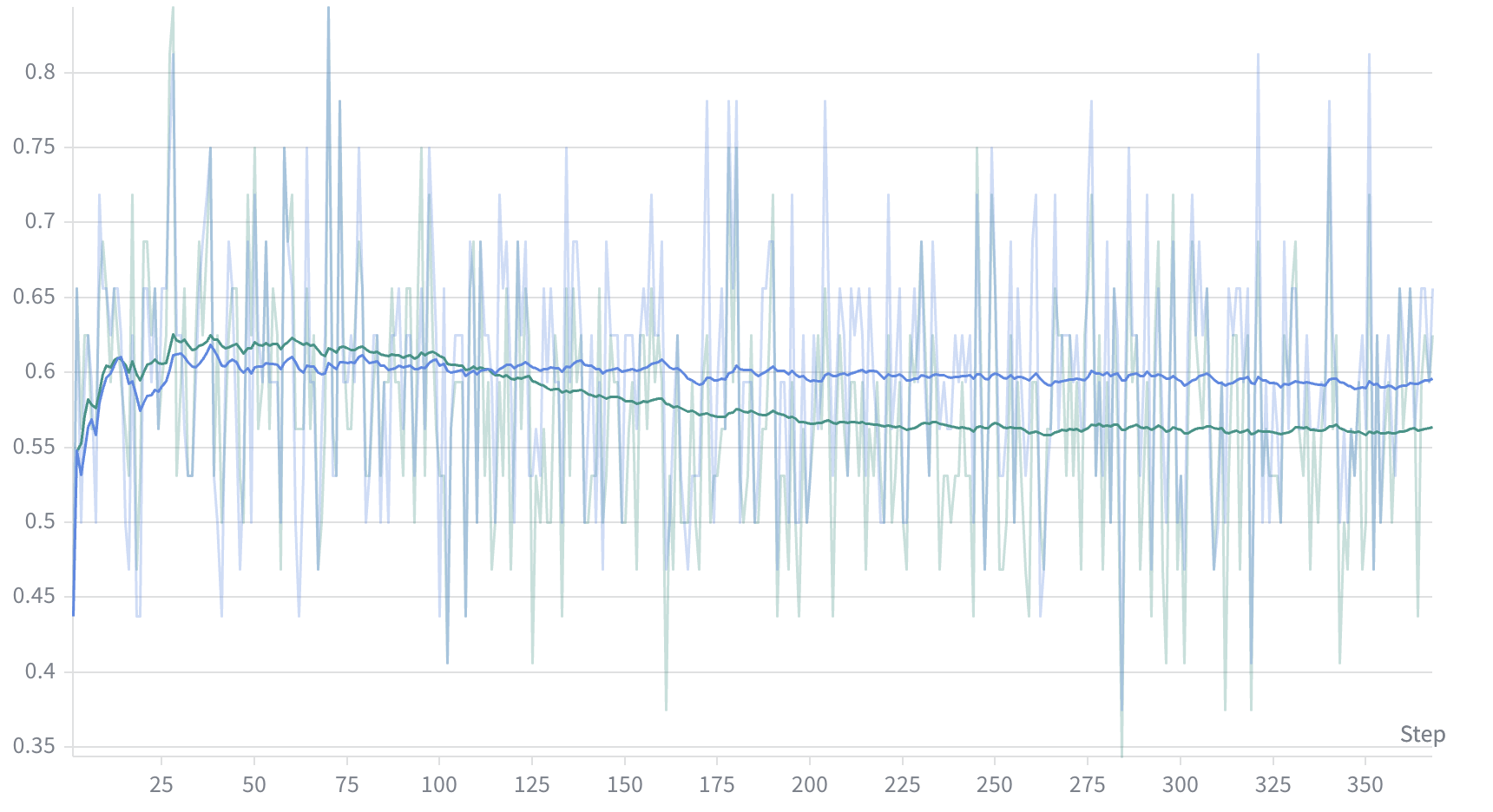}
    \end{minipage}
    \hfill
    \begin{minipage}[t]{0.285\textwidth}
        \centering
        \textbf{(c) Validation Avg@32}\par
    \vspace{1.5mm}
        \includegraphics[
            width=\linewidth,
            trim=0 0 0 0,
            clip
        ]{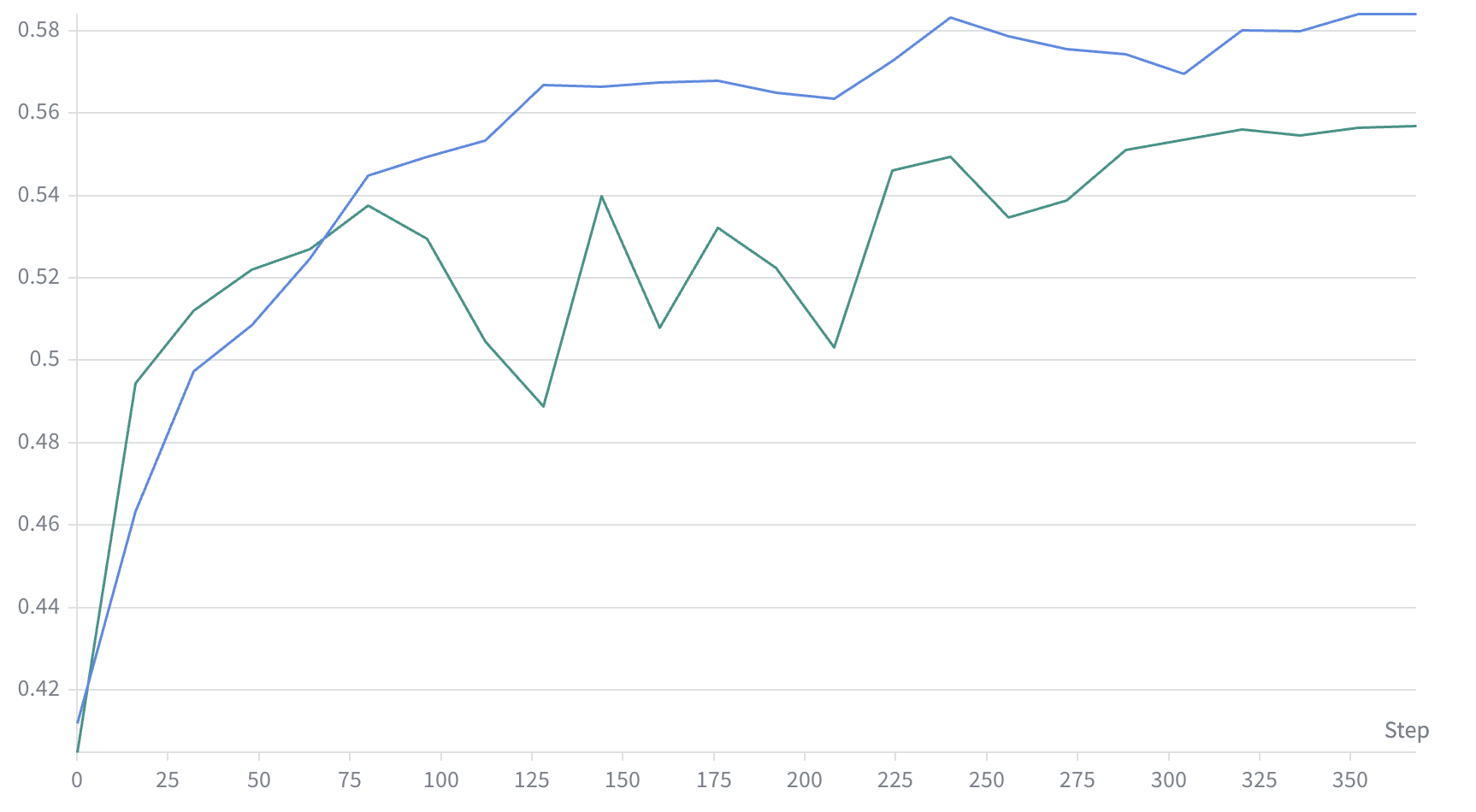}
    \end{minipage}

    \vspace{1mm}

    {\small
    \textcolor{blue}{\rule{9mm}{1.4pt}}\;With cache replay
    \qquad
    \textcolor{teal}{\rule{9mm}{1.4pt}}\;Without cache replay
    }

    \caption{
    Effect of cache replay during adaptation of
    Llama-3.2-3B-Instruct on MATH500.
    \textbf{Left:} cache-replay dynamics, including the cumulative number of
    prompts whose solutions have been cached and the number of previously
    mastered prompts recalled at each adaptation step.
    \textbf{Middle:} pseudo-label accuracy with and without cache replay;
    faint curves show per-step values and darker curves show smoothed
    trajectories.
    \textbf{Right:} validation Avg@32 accuracy throughout adaptation.
    Cache replay periodically revisits earlier self-consistent solutions,
    reducing degradation in pseudo-label quality and mitigating
    self-supervised forgetting. Removing cache replay lowers final Avg@32
    accuracy from $58.4\%$ to $56.3\%$.
    }
    \label{fig:cache-replay-analysis}
\end{figure*}

Figure~\ref{fig:cache-replay-analysis} examines the cache-replay mechanism
using Llama-3.2-3B-Instruct on MATH500. During adaptation, the cache
accumulates preference groups associated with previously mastered prompts
and periodically replays them in subsequent updates. As shown in
Figure~\ref{fig:cache-replay-analysis}(a), replay events revisit a subset of
these earlier solutions throughout training.

Figures~\ref{fig:cache-replay-analysis}(b)--(c) compare adaptation with and
without cache replay. Without replay, pseudo-label accuracy gradually
decreases and validation performance exhibits larger regressions. Replaying
cached preference groups anchors the policy to earlier self-consistent
solutions, improving the retention of previously acquired behavior and
raising final Avg@32 accuracy from $56.3\%$ to $58.4\%$. These results
support the use of cache replay as a mechanism for mitigating
self-supervised forgetting during online adaptation.

\subsection{Additional Adaptation Dynamics on MATH500}
\label{app:additional-dynamics}

\begin{figure*}[t]
    \centering

    \textbf{(a) Llama-3.2-3B-Instruct}\\[-1mm]
    \includegraphics[
        width=\textwidth,
        trim=0 0 0 0,
        clip
    ]{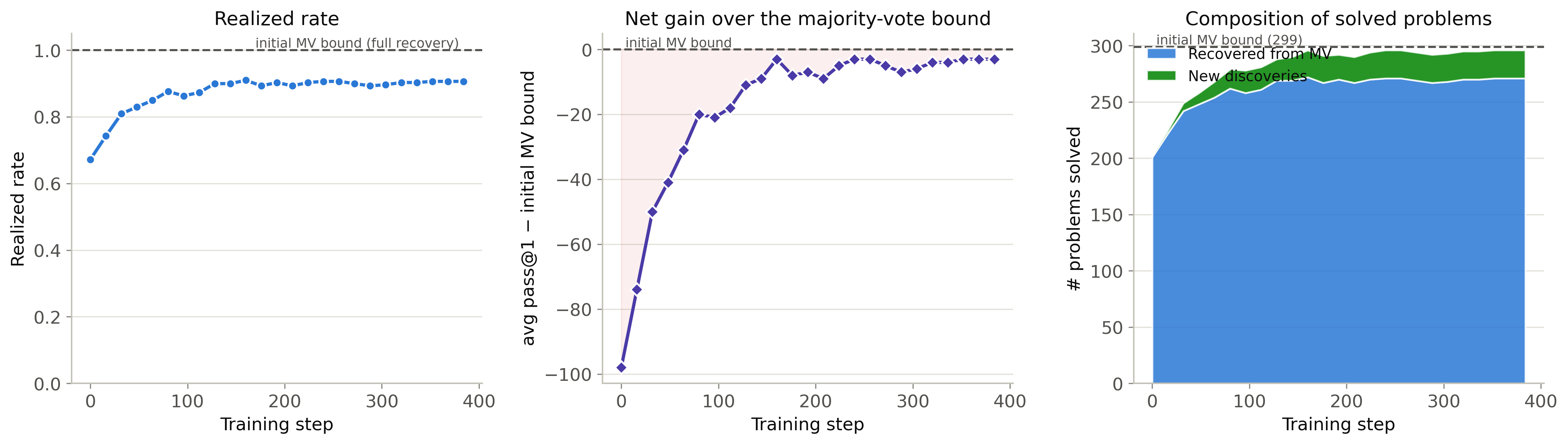}

    \vspace{2mm}

    \textbf{(b) Qwen2.5-Math-1.5B-Instruct}\\[-1mm]
    \includegraphics[
        width=\textwidth,
        trim=0 0 0 0,
        clip
    ]{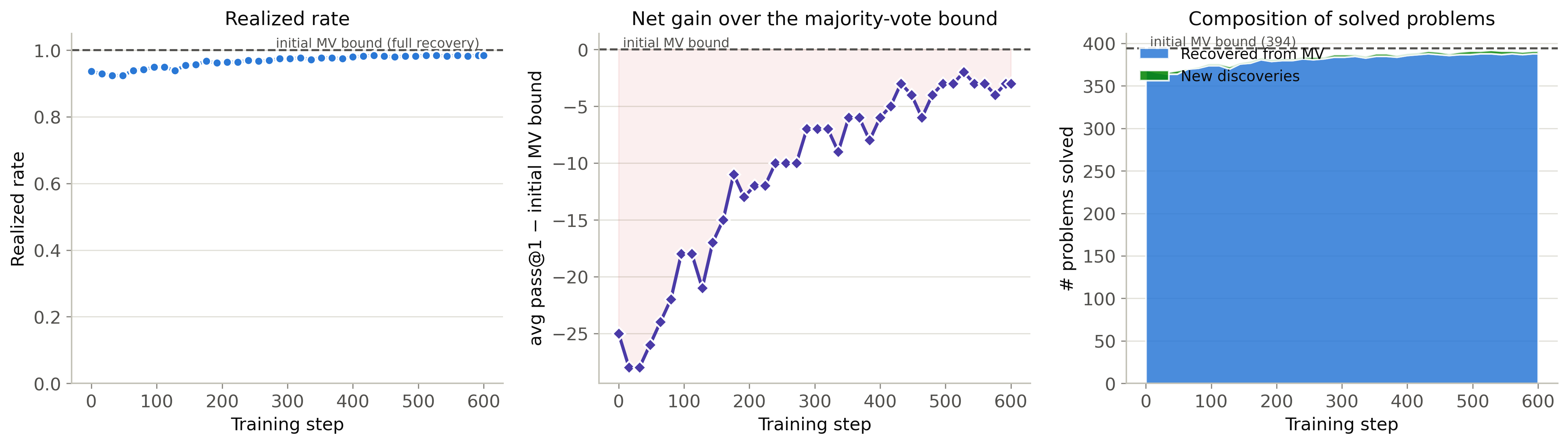}

    \caption{
    Additional adaptation dynamics on MATH500 for
    Llama-3.2-3B-Instruct and Qwen2.5-Math-1.5B-Instruct. In each row, the
    problems solved by the initial rollout majority define a fixed reference
    set.
    \textbf{Left:} fraction of this initial set recovered by the current
    policy.
    \textbf{Middle:} difference between the number of problems solved by the
    current policy and the initial majority-vote reference; the dashed line at
    zero denotes full numerical recovery of the initial reference level.
    \textbf{Right:} decomposition of the currently solved problems into
    solutions recovered from the initial reference set and newly discovered
    solutions. Both backbones recover most initially available solutions and occasionally
solve problems outside the initial reference set, while incomplete recovery
indicates degradation induced by systematic pseudo-label errors.
    }
    \label{fig:math500-dynamics}
\end{figure*}

Figure~\ref{fig:math500-dynamics} presents additional adaptation trajectories
on MATH500 for Llama-3.2-3B-Instruct and
Qwen2.5-Math-1.5B-Instruct. For each run, the problems solved by the initial
rollout majority define a fixed reference set. The left panels report the
fraction of this set recovered by the current policy, the middle panels show
the difference between the current number of solved problems and the initial
majority-vote reference, and the right panels decompose the currently solved
problems into recovered and newly discovered solutions.

Across both backbones, adaptation progressively recovers most solutions
available in the initial reference set. Llama-3.2-3B-Instruct recovers
approximately $90\%$ of the initially solvable problems, whereas
Qwen2.5-Math-1.5B-Instruct approaches nearly complete recovery. Both models
also solve some problems outside the initial reference set, showing that the
initial majority-vote-solvable set is not a strict performance upper bound.
Nevertheless, incomplete recovery remains, indicating that some initially
available solutions are lost during adaptation. These trajectories illustrate the competing effects of solution recovery,
cross-prompt transfer, and degradation induced by systematic pseudo-label
errors, which can steer policy updates away from previously available
solutions.

\end{document}